\documentclass[lettersize,journal,caption=false]{IEEEtran}
\IEEEoverridecommandlockouts
\usepackage{graphicx}
\def\BibTeX{{\rm B\kern-.05em{\sc i\kern-.025em b}\kern-.08em
    T\kern-.1667em\lower.7ex\hbox{E}\kern-.125emX}}
\usepackage[font=small,labelfont=bf,tableposition=top,hypcap=false]{caption}
\usepackage{blindtext}

\let\oldtwocolumn\twocolumn
\renewcommand\twocolumn[1][]{%
    \oldtwocolumn[{#1}{
    \begin{center}
    \vskip-5ex
        \centering
        \includegraphics[width=1.0\textwidth]{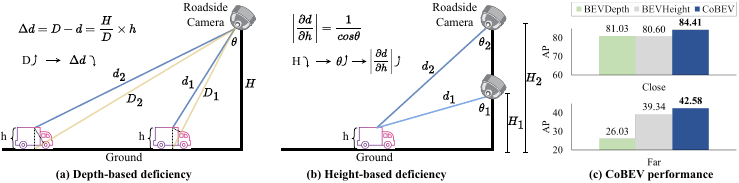}
        \captionof{figure} {{(a) As the distance ($d$) between the vehicle and the roadside camera increases, the depth difference (${\Delta}d$) between the vehicle and the ground decreases, leading to the unreliability of the depth-based lifting acquisition of the BEV features.
        (b) As the camera's height ($H$) decreases, the depth's partial differential with respect to the height ($\left|\frac{\partial d}{\partial h} \right|$) of the vehicle increases, leading to the unreliability of the height-based BEV features.
        (c) Vehicle monocular 3D detection results are presented for both close (range interval ${\sim}20m$) and far (range interval ${\sim}100m$) scenarios on the DAIR-V2X-I validation set~\cite{yu2022dair}. Our CoBEV leverages the fusion of complementary BEV features derived from both height and depth information, consistently achieving state-of-the-art performance across target-camera distances.
        }}
        \label{fig:teaser}
    \end{center}
    }]
}

\usepackage{amsmath,amsfonts}
\usepackage{algorithmic}
\usepackage{array}
\usepackage[caption=false,font=normalsize,labelfont=sf,textfont=sf]{subfig}
\usepackage{textcomp}
\usepackage{stfloats}
\usepackage{url}
\usepackage{verbatim}
\usepackage{graphicx}
\def\BibTeX{{\rm B\kern-.05em{\sc i\kern-.025em b}\kern-.08em
    T\kern-.1667em\lower.7ex\hbox{E}\kern-.125emX}}
\usepackage{balance}
\usepackage{booktabs}

\usepackage{multirow}
\usepackage[table]{xcolor}      
\usepackage{arydshln}           
\usepackage{bbding}            

\usepackage{xcolor}
\definecolor{rblue}{rgb}{0,0.5,1}
\definecolor{mask_red}{rgb}{1,0,0.8}
\newcommand{\green}[1]{\textcolor[RGB]{96,177,87}{#1}}

\newcommand{\red}[1]{\textcolor{mask_red}{#1}}
\usepackage{hyperref}
\hypersetup{colorlinks=true,linkcolor=red,citecolor=rblue}

\begin{document}

\title{CoBEV: Elevating Roadside 3D Object Detection with Depth and Height Complementarity}

\author{Hao~Shi\IEEEauthorrefmark{2},
        Chengshan~Pang\IEEEauthorrefmark{2},
        Jiaming~Zhang\IEEEauthorrefmark{2},
        Kailun~Yang\IEEEauthorrefmark{1},
        Yuhao~Wu,
        Huajian~Ni,
        Yining~Lin,
        Rainer~Stiefelhagen,
        and~Kaiwei~Wang\IEEEauthorrefmark{1}%
\thanks{This work was supported in part by the National Natural Science Foundation of China (NSFC) under Grant No. 12174341 and No. 62473139, in part by Shanghai SUPREMIND Technology Company Ltd., and in part by Hangzhou SurImage Technology Company Ltd.}%
\thanks{H. Shi and K. Wang are with the State Key Laboratory of Extreme Photonics and Instrumentation, Zhejiang University, Hangzhou 310027, China (email: haoshi@zju.edu.cn; wangkaiwei@zju.edu.cn).}%
\thanks{J. Zhang and R. Stiefelhagen are with the Institute for Anthropomatics and Robotics, Karlsruhe Institute of Technology, 76131 Karlsruhe, Germany (email: jiaming.zhang@kit.edu; rainer.stiefelhagen@kit.edu).}%
\thanks{J. Zhang was also with the Department of Engineering Science, University of Oxford, Oxford OX1 3PJ, UK.}%
\thanks{K. Yang is with the School of Robotics, Hunan University, Changsha 410012, China (email: kailun.yang@hnu.edu.cn).}%
\thanks{K. Yang is also with the National Engineering Research Center of Robot Visual Perception and Control Technology, Hunan University, Changsha 410082, China.}%
\thanks{H. Shi, C. Pang, Y. Wu, H. Ni, and Y. Lin are with Shanghai SUPREMIND Technology Co., Ltd, Shanghai 201210, China.}%
\thanks{\IEEEauthorrefmark{1}corresponding authors: Kaiwei Wang and Kailun Yang.}%
\thanks{\IEEEauthorrefmark{2}denotes equal contribution.}%
}

\markboth{IEEE Transactions on Image Processing,~September~2024}%
{Shi \MakeLowercase{\textit{et al.}}: CoBEV}
\maketitle

\begin{abstract}
Roadside camera-driven 3D object detection is a crucial task in intelligent transportation systems, which extends the perception range beyond the limitations of vision-centric vehicles and enhances road safety. While previous studies have limitations in using only depth or height information, we find both depth and height matter and they are in fact complementary. The depth feature encompasses precise geometric cues, whereas the height feature is primarily focused on distinguishing between various categories of height intervals, essentially providing semantic context. This insight motivates the development of Complementary-BEV (CoBEV), a novel end-to-end monocular 3D object detection framework that integrates depth and height to construct robust BEV representations. In essence, CoBEV estimates each pixel's depth and height distribution and lifts the camera features into 3D space for lateral fusion using the newly proposed two-stage complementary feature selection (CFS) module. A BEV feature distillation framework is also seamlessly integrated to further enhance the detection accuracy from the prior knowledge of the fusion-modal CoBEV teacher. We conduct extensive experiments on the public 3D detection benchmarks of roadside camera-based DAIR-V2X-I and Rope3D, as well as the private Supremind-Road dataset, demonstrating that CoBEV not only achieves the accuracy of the new state-of-the-art, but also significantly advances the robustness of previous methods in challenging long-distance scenarios and noisy camera disturbance, and enhances generalization by a large margin in heterologous settings with drastic changes in scene and camera parameters. For the first time, the vehicle AP score of a camera model reaches $80\%$ on DAIR-V2X-I in terms of easy mode. The source code will be made publicly available at \href{https://github.com/MasterHow/CoBEV}{CoBEV}.

\end{abstract}


\section{Introduction}

\IEEEPARstart{V}{ehicle-centric} 3D object detection has made significant strides in recent years~\cite{philion2020lift,li2022bevformer,li2022bevdepth}. 
{However, it continues to face substantial challenges related to occlusion in blind spots and constrained long-distance perception capabilities, stemming from the restricted observation of sensors mounted on vehicles.}
Conversely, infrastructure-centric 3D object detection exhibits a natural ability to capture long-range information, and is less susceptible to occlusion by vehicles, making it a crucial technology for the realization of safer and smarter transportation systems.
While LiDAR-based detectors located on the vehicle side continue to outperform camera-based detectors in terms of accuracy performance, 
camera-based methods have gained increased attention in roadside perception schemes due to their higher flexibility 
to adapt to existing traffic infrastructure, their lower cost, and ease of maintenance, compared to the expensive LiDAR alternatives~\cite{ye2022rope3d,yang2023bevheight}. 

In camera-based methods, the Bird's-Eye-View (BEV) space is extensively utilized because it not only encodes rich spatial context information but also provides a unified global-view feature space that is advantageous for the integration of various perception tasks such as
3D object detection~\cite{xie2023x,liu2022fine} and
semantic segmentation~\cite{pan2020cross,li2023bi},
mapping~\cite{chen2023trans4map,teng2023360bev}, \textit{etc.}
The BEV space also facilitates the promotion of Vehicle-Infrastructure Cooperative Autonomous Driving (VICAD)~\cite{yu2022dair, li2022v2x}.
As shown in previous work~\cite{philion2020lift,li2022bevformer,li2022bevdepth}, the primary challenge for BEV detection is how to ``lift'' the perspective data of the camera to the BEV space.
Previous state-of-the-art detection schemes typically rely on explicitly estimating the depth or height distribution of pixels~\cite{li2022bevdepth,yang2023bevheight} and utilizing camera parameters to perform feature projection.

While vehicle-side perception has made remarkable progress recently~\cite{li2022bevformer,li2022bevdepth}, roadside camera-based detectors face a unique set of challenges:
(1) Roadside cameras typically operate individually with a monocular setup, making it difficult to integrate multi-view images to compute accurate depth.
(2) The observation distance of the roadside camera is much longer than the vehicle view, as depicted in Fig.~\ref{fig:teaser}(a), making it difficult for a depth-based detector that operates effectively at the vehicle end, to distinguish between distant vehicles and road surface features.
(3) The camera parameters of the roadside camera can be various.
{For instance, while vehicle-mounted cameras are typically aligned with the direction of travel and fixed at a uniform height, roadside cameras at different intersections exhibit wide variations in their extrinsic parameters.}
As shown in Fig.~\ref{fig:teaser}(b), we observe that the differential of depth calculated from height increases for previous height-based detectors, as the camera mounting height decreases, implying that inaccurate height estimates will introduce larger errors in object detection.

On the other hand, we find that the features of depth and height detectors exhibit complementary properties. 
As shown in Fig.~\ref{fig:teaser}(c), in long-distance scenes that are challenging for depth-based methods, it is more precise to regress height to calculate BEV features, due to the constant height of vehicles.
Conversely, when the distance is closer, directly regressing depth becomes comparatively easier. In terms of feature encoding, we argue that depth detectors rely more on precise geometric cues, whereas height detectors learn distinct height distribution intervals across different categories, thus placing greater emphasis on the semantic context of targets. We substantiate this observation through ablation in Tab.~\ref{tab:pcd-supervision}.

Based on the above observations, we hereby propose \textbf{Co}mplementary-\textbf{BEV} (\textbf{CoBEV}), 
a roadside monocular 3D object detection framework that seamlessly integrates the complementary depth and height cues to establish robust BEV representations for elevating traffic scene understanding. 
Specifically, we introduce a Camera-aware Hybrid Lifting (CHL) module (Sec.~\ref{sec:camera-aware_hybrid_lifting}) to independently estimate the depth and height distribution for each pixel. This module lifts camera features into the BEV space by deriving a unified context. Subsequently, the heterogeneous 3D features undergo partial-pillar voxel pooling, reducing computational complexity while preserving the free flow of information in the height dimension. These condensed heterogeneous features are then put into the novel Complementary Feature Selection (CFS) module (Sec.~\ref{sec:complementary_feature_selection_module}), where global pattern and local cues are selectively fused to generate a robust BEV representation. Furthermore, we present a new BEV Feature Distillation framework (Sec.~\ref{sec:bev_feature_distillation}) capable of enhancing detection accuracy agnostic to various target sizes, without introducing additional computational complexity.
Thanks to these core designs, our CoBEV achieves state-of-the-art performance, as shown in Fig.~\ref{fig:teaser}(c). Considering the diversity and variability of the infrastructure scenarios, generalization ability is crucial for the improvement of intelligent transportation systems. Therefore, we first present a thorough analysis of how the BEV representation quality affects the mono3D object detection's robustness and generalization across heterologous roadside datasets and camera parameters disturbance. 

To validate the power of the proposed CoBEV, we conduct extensive experiments on two public roadside datasets DAIR-V2X-I~\cite{yu2022dair} and Rope3D~\cite{ye2022rope3d}, and a private dataset Supremind-Road.
CoBEV achieves the new state-of-the-art performance of $69.57\%$ / $47.21\%$ / $66.17\%$ in 3D Average Precision ($AP_{3D|R40}$) across Vehicle, Pedestrian and Cyclist on the DAIR-V2X-I validation set, outperforms the second-best BEVHeight~\cite{yang2023bevheight} with a large margin of $3.80\%$ / $7.92\%$ / $6.09\%$, respectively. On the Rope3D, CoBEV surpasses all other methods with $AP_{3D|R40}$ of $52.72\%$ / $29.28\%$ for Car and Big Vehicle detection under the difficult $IoU{=}0.7$ setting, perform a significant improvement of $6.99\%$ / $6.21\%$ compared with the previous best-published method.  
{In practical settings, camera parameters and visual capture vary significantly across intersections, making it infeasible to collect and annotate data for every intersection in each city.}
Therefore, we also compare the robustness of different BEV representations by studying heterologous settings and camera parameters disturbance and observe a significant drop in the accuracy of the previous state-of-the-art approach.
Compared with contemporary methods~\cite{yang2023bevheight}, CoBEV achieves an average $1.61\%$ improvement on the challenge Rope3D heterologous setting across all categories, and $2.56\%$ improvement on Supremind-Road heterologous setting, thanks to the complementary robust BEV representations, which further demonstrates the effectiveness of the proposed CoBEV solution. In scenarios where camera parameters are directly perturbed by noise affecting focal length, roll, and pitch, CoBEV exhibits an enhanced detection capability, surpassing BEVHeight by an average of $3.93\%$. This once again demonstrates CoBEV's notable resistance to interference and robustness.

In summary, we deliver the following contributions:
\begin{itemize}
    \item We present a comprehensive theoretical examination of the shortcomings of prior approaches that exclusively rely on either depth or height
    for constructing BEV features in infrastructure-centric environments.
    \item We propose CoBEV, a novel roadside monocular 3D detection framework that generates robust BEV representations by seamlessly incorporating complementary geometry-centric depth and semantic-centric height cues.
    \item An in-depth study is conducted to examine the impacts of various feature fusion methods on constructing robust BEV features through extensive ablation experiments.
    \item The superior generalization ability of CoBEV in diverse and heterologous settings, as well as its performance on noisy camera parameters, proves  
    its suitability for roadside scenarios for elevating traffic scene understanding.
    \item Extensive experiments on two public and one private dataset demonstrate that CoBEV achieves state-of-the-art detection accuracy, establishing it as a robust and reliable solution for roadside 3D detection.
\end{itemize}

\section{Related Work}
\subsection{Camera-based 3D Object Detection}
{3D object detection typically utilizes inputs from LiDAR or cameras, where LiDAR-based methods are generally categorized into point cloud-based~\cite{shi2019pointrcnn,yang20203dssd,zhang2022not} and grid-based~\cite{chen2017multi,zhou2018voxelnet,yan2018second} paradigms. In this paper, we focus on monocular 3D detection using camera-based methods.}
Camera-based 3D object detection aims to predict the 3D bounding box of the object from an image. According to the deployment scenarios, they can be divided into two types: vehicle-centric and infrastructure-centric methods.

\noindent \textbf{Vehicle-centric methods.}
Vehicle-centric methods have been extensively explored in previous works.
One branch of methods directly uses 2D detectors and slightly modifies the detection head to achieve 3D detection~\cite{duan2019centernet,tian2019fcos,zhu2020deformable}. FCOS3D~\cite{wang2021fcos3d} adapts 2D detectors by predicting both 2D and 3D attributes.
{DETR3D~\cite{wang2022detr3d} introduces 3D object queries for direct regression of 3D bounding boxes. Extending this approach, PETR~\cite{liu2022petr} incorporates 3D position-aware representation to enhance accuracy. PETRv2~\cite{liu2023petrv2} further extends this by incorporating temporal information through enhanced 3D position embedding. StreamPETR~\cite{wang2023exploring} introduces object-centric queries to aggregate long-term historical temporal features. Far3D~\cite{jiang2024far3d} employs a DepthNet and a 2D detector to initially predict 3D queries and then refines them for precise 3D bounding box predictions.}
Another branch of vehicle-centric methods performs detection directly in 3D feature space.
OFT~\cite{roddick2018orthographic} is the first to use the orthographic feature transform to transform 2D image features into 3D space for monocular 3D detection. LSS~\cite{philion2020lift} achieves image-based 3D detection by regressing the depth distribution to transform image features into BEV space. ImVoxelNet~\cite{rukhovich2022imvoxelnet} projects a voxel grid into the image feature space to generate a voxel representation.
BEVDepth~\cite{li2022bevdepth} introduces the depth of LiDAR for supervision on the basis of LSS, which improves the accuracy of depth estimation and achieves state-of-the-art detection accuracy.
CrossDTR~\cite{tseng2022crossdtr} builds depth-aware embedding from depth estimation to improve performance using transformer backbones.
Different from vehicle-centric works, we focus on infrastructure-centric roadside-camera-driven 3D object detection for elevating traffic scene understanding.

\noindent \textbf{Infrastructure-centric methods.}
While roadside devices have a wider range of perception than vehicle end and can achieve long-term observation, it has been under-explored in the literature. 
Recently, V2X-SIM~\cite{li2022v2x} is the first to use the CARLA simulator~\cite{dosovitskiy2017carla} to collect a public vehicle-road collaboration dataset.
DAIR-V2X~\cite{yu2022dair} and Rope3D~\cite{ye2022rope3d} have also released vehicle-road perception data in the real scene, establishing a benchmark for roadside detection.
Yet, the accuracy of state-of-the-art vehicle detection algorithms~\cite{li2022bevdepth} at the roadside is limited~\cite{yang2023bevheight}.
To address this issue, BEVHeight~\cite{yang2023bevheight} first focuses on roadside detection and proposes predicting the height distribution of the scene instead of the depth distribution, which improves the performance.
In contrast, CBR~\cite{fan2023calibration} maps image features to BEV space based on Multi-Layer Perceptrons (MLPs), bypassing the step of extrinsic calibration, but the accuracy is limited.
Different from these methods, we propose CoBEV, which fully considers the challenges of long-distance targets and diverse camera parameters unique to roadside 3D detection for elevating perception performance and robustness by seamlessly fusing geometry-centric depth and semantic-centric height cues.

{Notably, a \textit{concurrent} work, BEVHeight++~\cite{yang2023bevheight++}, also explores the fusion of depth and height BEV features to further enhance detection accuracy. This trend underscores the evolving landscape of infrastructure-centric detection. While CoBEV shares similarities with BEVHeight++, including their focus on integrating depth- and height-based BEV features, several technical differences distinguish them: First, CoBEV empirically demonstrates the geometric importance of depth features and the semantic significance of height features. 
{Moreover, CoBEV employs partial-pillar voxel pooling to preserve the vertical axis and enhance free-flowing feature fusion, while BEVHeight++ uses deformable transformer-based attention for fusion.}
Furthermore, CoBEV introduces a BEV feature distillation framework to learn robust BEV features from multi-modal detections. These distinctions highlight the unique contributions and potential advantages of the CoBEV framework in advancing roadside 3D perception.}

\subsection{BEV Scene Understanding}
The concept of Bird's-Eye-View (BEV) perception has gained increasing attention due to its ability to provide a unified feature space and rich spatial context, making it particularly well-suited for traffic scenarios.
A crucial question in this field is how to construct BEV features from the images captured by pinhole cameras. The existing methods can be categorized into two categories: implicit and explicit methods.

\noindent \textbf{Implicit lifting methods.}
The implicit methods for constructing BEV features utilize MLPs or transformers. 
The MLP-based approach offers a straightforward mapping strategy where image features are fully connected to BEV features.
VPN~\cite{pan2020cross} initially utilizes the global mapping capability of MLP to map the multi-view camera input to the top-down feature map for semantic segmentation.
Subsequent methods~\cite{hendy2020fishing,yang2021projecting} also employ MLP for view transformation to implement road layout estimation, semantic segmentation, \textit{etc.}
{In contrast, transformer-based methods define BEV grid queries or object queries and perform cross attention with image features to construct BEV representations~\cite{li2022bevformer,wang2022detr3d,liu2022petr,yan2023cross,yang2023bevformer}.}
The most well-known of these methods, BEVFormer~\cite{li2022bevformer}, employs spatial cross-attention for perspective transformation and considers feature aggregation based on temporal attention.
However, these implicit methods do not fully consider the corresponding camera parameters, which limits their generalization to diverse roadside scenarios.

\noindent \textbf{Explicit lifting method.}
The explicit approach for constructing BEV features is based on geometric principles.
The earliest Inverse Perspective Mapping (IPM)~\cite{mallot1991inverse} is a straightforward method for achieving BEV perception, which utilizes horizontal plane constraints and physical mapping for perspective transformation.
However, directly applying IPM transformation on images is hindered by sampling errors and quantization effects. 
On the other hand, a cluster of methods attempts to incorporate depth estimation to mitigate the ill-posed issues associated with monocular 3D perception.
The Pseudo LiDAR~\cite{wang2019pseudo} pioneers to employ off-the-shelf depth estimation techniques to convert image pixels into pseudo-LiDAR-point-clouds, which can then be processed in 3D space.
Recently, LSS~\cite{philion2020lift} stands out as the dominant paradigm, where depth or height distributions in 3D space are explicitly estimated and camera parameters are utilized for feature projection~\cite{li2022bevdepth,yang2023bevheight,li2023bi,chen2023trans4map,teng2023360bev}
Due to the incorporation of camera parameters, explicit methods are more suitable for BEV feature generation in roadside scenes.
In this work, depth and height are leveraged to achieve view transformation, and their distinctions and complementarities are thoroughly considered to extract robust BEV features from monocular cameras.
{Consequently, our approach enables satisfactory monocular 3D detection across diverse heterologous intersection data not included in the training set, while effectively handling noisy camera parameters.}

\section{CoBEV: Proposed Framework}
\begin{figure*}[!t]
\centering
\includegraphics[width=0.9\linewidth]{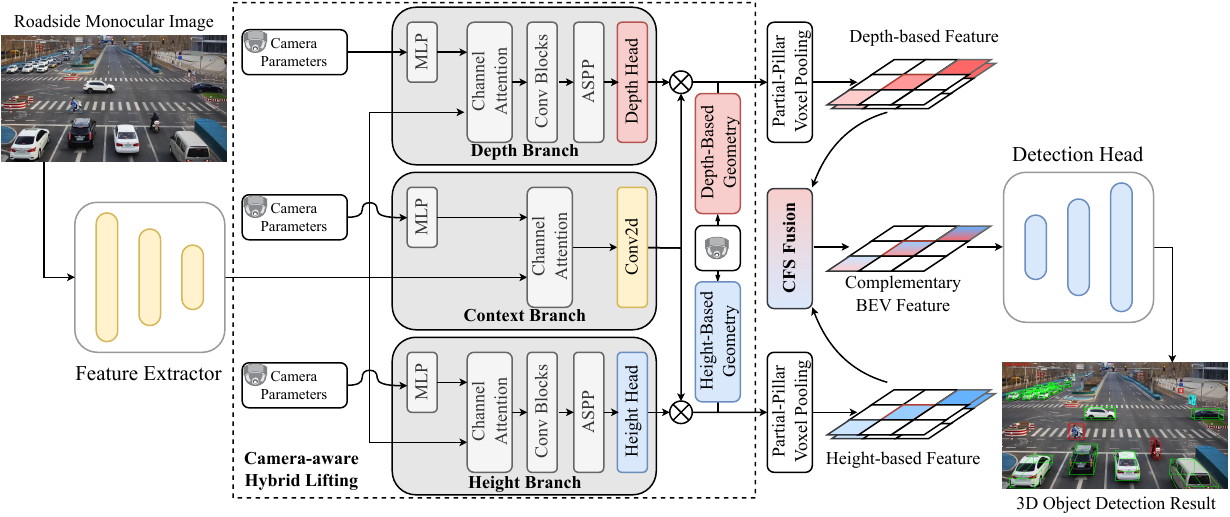}
\caption{{\textit{Overview of the Complementary-BEV ({CoBEV}) architecture.} 
Firstly, the monocular image on the roadside is fed into the feature extractor to encode high-dimensional features. Image features are then sent to the Camera-aware Hybrid Lifting (CHL) module that consists of a depth branch, a context branch, and a height branch, and fused with the camera parameters encoded by the MLP. The pixel distribution of the depth and height branches are integrated with the context feature via an outer product to obtain a frustum-shaped point cloud. The point cloud coordinates can be obtained by depth or height geometry (see Fig.~\ref{fig:lifting} for details). This point cloud is then splatted to the depth-based and height-based compressed 3D features by partial-pillar voxel pooling. Finally, the multi-source BEV features are fused via the Complementary Feature Selection (CFS) module to develop robust BEV features for 3D object detection. 
}
}
\vskip -3ex
\label{overview}
\end{figure*}

In this section, the problem formulation of the roadside monocular 3D object detection is first provided (Sec.~\ref{sec:problem_formulation}). 
Next, the core designs of CoBEV are described in detail, including the \textit{Camera-aware Hybrid Lifting} module (Sec.~\ref{sec:camera-aware_hybrid_lifting}), \textit{Complementary Feature Selection (CFS)} module (Sec.~\ref{sec:complementary_feature_selection_module}), and the \textit{BEV Feature Distillation} (Sec.~\ref{sec:bev_feature_distillation}).

\subsection{Problem Formulation}
\label{sec:problem_formulation}

In this work, we aim to build a robust roadside 3D monocular detector:
\begin{equation}
\label{equ:formulation}
    \mathbf{B_{ego}} = Det(X, E, I | \phi),
\end{equation}
where $Det(\cdot)$ is the detector model, $\phi$ is the learned parameters, $X {\in} \mathbb{R} ^ {3 \times H \times W}$ is the input single frame capture of the infrastructure camera, $E {\in} \mathbb{R} ^ {3 \times 4}$ and $I {\in} \mathbb{R} ^ {3 \times 3}$ are the extrinsic and intrinsic matrix of the image $X$, respectively. $H$ and $W$ denote the input image's height and width. 
$\mathbf{B_{ego}}$ is a series of output 3D bounding box:
\begin{equation}
\label{equ:b_ego}
    \mathbf{B_{ego}} = \{ \hat{B}_{1}, \hat{B}_{2}, ... , \hat{B}_{n} \},
\end{equation}
where $n$ is the number of foreground targets and $\hat{B}_{i}$ is a vector with $7$ degrees of prospects, which are composed of position, dimension, and orientation:
\begin{equation}
\label{equ:b_i}
    \hat{B}_{i} = (x,y,z,l,w,h,\theta),
\end{equation}
where $(x,y,z)$ is the position of each 3D bounding box, $(l,w,h)$ is the three-dimensional size of its corresponding cuboid, and $\theta$ is the yaw angle relative to the ground plane.

For roadside cameras, the extrinsic and intrinsic parameters can be obtained through calibration. 
{Given data ownership among different camera devices and privacy regulations, our study concentrates on the 3D detection using single-frame monocular cameras, which are prevalent in intersection scenes and offer substantial practical value for intelligent transportation systems.}

\subsection{Camera-aware Hybrid Lifting}
\label{sec:camera-aware_hybrid_lifting}
Lifting 2D features to 3D space is crucial for BEV perception~\cite{philion2020lift, li2022bevdepth, yang2023bevheight}. 
As shown in Fig.~\ref{fig:teaser}, 
{
we illustrate the complementarity of depth and height detectors from the perspectives of mathematical derivation:
\noindent \textbf{a) Sensitivity analysis of the depth detector.}
For depth detectors, the semantic category judgment relies on the difference between the top and bottom depths of the target, $\Delta d{=}D{-}d$. As $D$ increases, $d$ approaches $D$:}
{
\begin{equation}
\lim_{D \to \infty} d = \lim_{D \to \infty} D \sqrt{1 - \frac{2Hh}{D^2} + \frac{h^2}{D^2}} \approx D.
\end{equation}}{\noindent Therefore, as $D$ increases, $\Delta d$ approaches $0$, making it more difficult for the depth detector to accurately determine the target category at long distances. This mathematical analysis aligns with the quantitative results in Fig.~\ref{fig:depth-ap}, which evaluate the long-range accuracy distribution of depth and height detectors. The height difference $h$ between the upper and lower bounds remains unchanged, and the impact of long-range on the height detector is smaller, compensating for the shortcomings of the depth detector.}
{
\textbf{b) Sensitivity analysis of the height detector.}
For height detectors, depth $d$ is obtained from the estimated height $h$ combined with camera parameters. The ratio $\frac{d}{h}$ reflects the impact of uncertainty in $h$ on $d$. By finding the partial differential of $\frac{d}{h}$ with respect to $H$, we can analyze this uncertainty:
}
{
\begin{equation}
\frac{\partial}{\partial H} \left( \frac{d}{h} \right) = -\frac{1}{h} \cdot \frac{1}{\sqrt{\left(\frac{D}{h}\right)^2 - \frac{2H}{h} + 1}}.
\end{equation}
}{\noindent The derivative is always negative, indicating that as the camera installation height $H$ decreases, $\frac{d}{h}$ increases. This shows that reducing $H$ increases the uncertainty in $d$ calculation due to $h$ estimation inaccuracies, making it harder for the height detector to place features accurately in 3D space. However, since $d$ also decreases with $H$, the uncertainty of the depth detector decreases, compensating for the height detector's shortcomings.
}
{Analysis of depth and height distributions on the roadside reveals that error increases with distance between the vehicle and the facility due to depth lifting, and depth estimation error rises as height decreases, attributable to height lifting.}
In other words, the geometric representations of depth and height are actually complementary in the transportation infrastructure scene as they are mutually beneficial in terms of accuracy.
This is attributed to the fact that 1) when depth is ambiguous, the height of the camera and the car remain constant, 2) as the camera height decreases, the depth itself decreases, making it easier to estimate the geometry directly, and, 3) depth information encodes precise geometric cues, whereas height distributions better capture distinct classes of semantic context.

Motivated by these key observations, we introduce
the \textit{Camera-aware Hybrid Lifting (CHL)} to obtain multi-source 3D features leveraging both depth and height information.
As shown in Fig.~\ref{overview}, this novel module is distinctly compartmentalized into three branches, including the depth branch, the context branch, and the height branch. 
Camera parameters $ \{ E,I \} $ are first encoded as camera-aware features $F_{cam}$ by three individual Multi-Layer Perceptron (MLP):
\begin{equation}
\label{equ:cam_mlp}
    F_{cam} = MLP( \xi (E) \oplus \xi (I)).
\end{equation}
To elucidate further, both the depth $D$ and height $H$ distributions involve a discretization process. The depth is discretized by performing uniform discretization (UD), which entails partitioning it into a pre-defined number of bins $N_{D}$ and the certain range of depth $(D_{0}{\sim}D_{max})$:
\begin{equation}
\label{equ:delta_d}
   \begin{split}
      \left\{
      \begin{aligned}
         \delta D & = \frac{D_{max} - D_{0}}{N_{D}}, \\
       D_{i} & = i \cdot \delta D,     
      \end{aligned}
      \right.
   \end{split}
\end{equation}
where the index $i{\in}\lbrack 1,N_{D} \rbrack$. Considering the variability in camera height fluctuations observed within different infrastructure facilities, the number of height bins $N_{H}$ is also fixed, yet the height range represented by each bin is subject to dynamic adjustments in accordance with the specific scene:
\begin{equation}
\label{equ:delta_h}
       H_{j} = \lfloor H_{0} + (\frac{j}{N_{H}})^{\alpha} \cdot (H_{max}-H_{0}) \rfloor,     
\end{equation}
where the index $j {\in} \lbrack 1, N_{H} \rbrack$, $\alpha$ denotes the hyperparameter that controls the dynamic expansion of each height bin.
We empirically set $\alpha{=}1.5$ in all experiments.

Given the inherent similarity of the regression tasks, the depth branch and the height branch exhibit a completely symmetric structure, and the prediction of depth and height distribution can be represented via the following formula:
\begin{equation}
\label{equ:depth_height}
   \begin{split}
      \left\{
      \begin{aligned}
        D_{i}^{pred} & = & \psi_{d}(CA_{d}(F|F_{cam})), \\
        H_{j}^{pred} & = & \psi_{h}(CA_{h}(F|F_{cam})),     
      \end{aligned}
      \right.
   \end{split}
\end{equation}
where $\psi_{d}(\cdot)$ and $\psi_{h}(\cdot)$ represent the height- and the depth network, respectively. $CA(\cdot)$ denotes the channel attention layer. $D_{i}^{pred} {\in} \mathbb{R} ^ {N_{D} \times \frac{H}{16} \times \frac{W}{16}}$ and $H_{j}^{pred} {\in} \mathbb{R} ^ {N_{H} \times \frac{H}{16} \times \frac{W}{16}}$ are the predicted depth and height distribution along the entire ray of viewpoint.

Upon obtaining the pixel-wise distributions of depth and height, the subsequent step in our methodology involves the projection of image features into the volumetric space of the encompassing frustum-shaped structure.
To commence this process, we initiate with the computation of spatial context information $F_{context}$ derived from the image features $F$, which can be represented as follows:
\begin{equation}
\label{equ:context}
      F_{context} = Conv_{2d}(CA_{c}(F|F_{cam})),     
\end{equation}
where $Conv_{2d}(\cdot)$ denotes the 2D convolutional block.
Subsequently, we employ the outer product operation to derive the three-dimensional feature representations, yielding:
\begin{equation}
\label{equ:3d_feat}
   \begin{split}
      \left\{
      \begin{aligned}
        & F_{depth}^{3d} & = F_{context} \otimes D_{pred}, \\
        & F_{height}^{3d} & = F_{context} \otimes H_{pred},     
      \end{aligned}
      \right.
   \end{split}   
\end{equation}
where $\otimes$ represents the outer product operation.
This intricate procedure facilitates the transformation of image-based data into a comprehensive 3D context, thereby enhancing the depth and height information integration within our framework.

\begin{figure}[!t]
\centering
\includegraphics[width=0.7\linewidth]{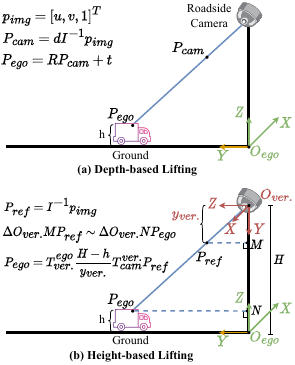}
\caption{\textit{Camera-based Hybrid Lifting} includes
(a) Explicit lifting based on the depth distribution and camera parameters.
(b) Explicit lifting with similar triangles based on the height distribution and camera parameters.
}
\label{fig:lifting}
\vskip -4ex
\end{figure}

Subsequently, the two heterologous three-dimensional volume $F_{depth}^{3d} {\in} \mathbb{R} ^ {N_{D} \times C \times \frac{H}{16} \times \frac{W}{16}}$ and $F_{height}^{3d} {\in} \mathbb{R} ^ {N_{H} \times C \times \frac{H}{16} \times \frac{W}{16}}$ undergo a meticulous transformation into the unified ego coordinate system.
This ego coordinate system is meticulously centered on the ground plane, a visual representation of depth-based and height-based coordinate transformation is illustrated in Fig.~\ref{fig:lifting}.
For the transformation of depth-based features, as shown in Fig.~\ref{fig:lifting}(a), a direct conversion is executed utilizing both intrinsic and extrinsic matrices:
\begin{equation}
\label{equ:depth_transform}
      P_{ego}^{depth}=R I^{-1} \lbrack ud,vd,d \rbrack^{T} + t,   
\end{equation}
where each 2D point on the image plane is represented with $p_{img}{=}\lbrack u,v,1 \rbrack^{T}$, $u$ and $v$ are the pixel index. $d$ denotes the corresponding depth. $R$ and $t$ is the rotation matrix and translation matrix with $E{=}\lbrack R,t \rbrack$.
In contrast, when handling height-based features, a two-step process is implemented as shown in Fig.~\ref{fig:lifting}(b). Initially, pixels $p_{img}$ from the image $I$ are transformed into the normalized coordinate system of the camera, \textit{i.e.}, the reference plane:
\begin{equation}
\label{equ:height_transform_1}
      P_{ref}^{height}=I^{-1} \lbrack u,v,1 \rbrack^{T}.     
\end{equation}
Subsequently, a vertical coordinate system is employed to compute the indices of the ego point $P_{ego}$, whose origin $O_{ver.}$ is the camera optical center, and its $Y$ axis is perpendicular to the ground. This calculation is performed by constructing a pair of similar triangles:
\begin{equation}
\label{equ:height_transform_2}
      \Delta O_{ver.}MP_{ref} \sim \Delta O_{ver.}NP_{ego}.   
\end{equation}
Concretely, the ego point can be obtained by:
\begin{equation}
\label{equ:height_transform_3}
      P_{ego}^{height}=T^{ego}_{ver.} \frac{H-h}{y_{ver.}} T^{ver.}_{cam} I^{-1} \lbrack u,v,1 \rbrack^{T},     
\end{equation}
where $T^{ver.}_{cam}{\in}\mathbb{SE}(3)$ is the transformation matrix from camera coordinate to vertical coordinate. 
$T^{ego}_{ver.} {\in} \mathbb{SE}(3) $ denotes the transformation matrix from the vertical coordinate to the uniformed ego coordinate system. $y_{ver.}$ represents the distance between $P_{ref}$ and camera center point $O_{ver.}$ along the vertical coordinate's $Y$ axis.
This rigorous transformation process serves to align the volumetric data within a consistent ego coordinate system, thereby facilitating subsequent perception in a unified manifold.

\subsection{Complementary Feature Selection} 
\label{sec:complementary_feature_selection_module}
After the lifting process, two large 3D point clouds unified in the ego coordinate system can be obtained.
To save subsequent calculations, a partial-pillar voxel pooling operation is performed on the point cloud. 
In the context of PointPillars~\cite{lang2019pointpillars}, a `pillar' refers to an infinite height voxel grid. 
{Unlike previous methods~\cite{philion2020lift,li2022bevdepth,yang2023bevheight}, our approach segments each pillar into multiple partial-pillar-shaped voxels, thus retaining the height dimension after voxel pooling instead of discarding it:}
\begin{equation}
\label{equ:partial_pooling_v2}
      \hat{F}^{3d}(\hat{p},i,j,k) = \sum\limits_{p=r \cdot \hat{p}}^{r \cdot (\hat{p}+1)}{F^{3d}(p,i,j,k)},   
\end{equation}
where $F^{3d} {\in} \mathbb{R} ^ {D \times C \times \frac{H}{16} \times \frac{W}{16}}$ denotes the large 3D point cloud. 
$\hat{F}^{3d}{\in}\mathbb{R} ^ {\frac{D}{r} \times C \times \frac{H}{16} \times \frac{W}{16}}$ represents the output compressed feature with reduction factor $r$. This design aims to ensure that visual cues related to detection can freely flow in all dimensions during the subsequent feature selection process, thereby achieving task-centric fusion features. Furthermore, preserving the height axis also mitigates the information bottleneck before and after fusion operations.

Next, the Complementary Feature Selection (CFS) module will be discussed in detail. The key idea is to select and construct the most relevant BEV features for the detection task from two heterogeneous compressed 3D features.

\begin{figure}[!t]
\centering
\includegraphics[width=0.8\linewidth]{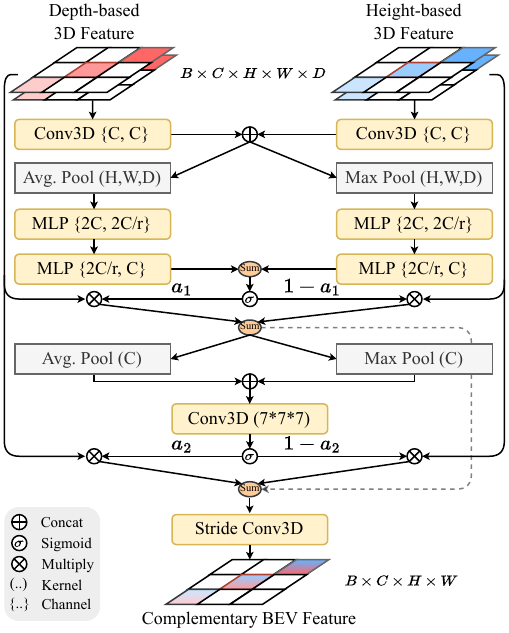}
\caption{\textit{The proposed Complementary Feature Selection (CFS) module.} 
Different from previous works~\cite{yang2023bevheight}, we maintain a low-scale vertical axis of depth and height BEV features after voxel pooling to promote information flow in the feature fusion process. CFS consists of two cascade feature selection processes, with the first stage for selecting complementary features in the column-shape channels, and the second stage for selecting features in the BEV plane, which are ultimately compressed to two-dimensional complementary BEV features through the stride 3D convolutional compression.
}
\label{fig:fusion}
\vskip -4ex
\end{figure}

As shown in Fig.~\ref{fig:fusion}, in the first stage of CFS, we concatenate the depth and height 3D features in parallel, and activate the grid after spatial pooling to obtain the affinity of first-stage selection $a_{1}$ and complete the first feature selection:
\begin{equation}
\label{equ:cfs_g}
      g = \mathcal{W}_{2}(\mathcal{W}_{1}(\mathcal{P}_{3d}(\hat{F}^{3d}_{depth} \oplus \hat{F}^{3d}_{height}))),     
\end{equation}
where $g$ denotes the global feature context. 
$\oplus$ is the concatenation. 
$\mathcal{P}_{3d}(\cdot)$ represents 3D global pooling layer. 
$\mathcal{W}_{1}$ is a dimension reduction layer, and $\mathcal{W}_{2}$ is a dimension increasing layer with the channel reduction ratio $r$, both of which are implemented with MLP. 
Then, we calculate the affinity $a_{1}$ of the first-stage feature selection:
\begin{equation}
\label{equ:cfs_a1}
      a_{1} = \sigma(g_a \uplus g_m),     
\end{equation}
where $\uplus$ denotes the element-wise summation, $\sigma(\cdot)$ is the $Sigmoid$ function. $g_a$ and $g_m$ are the global feature context from 3D average pooling branch and 3D max pooling branch, respectively. The first-stage selection feature $F^{1}_{s}$ can be obtained by:
\begin{equation}
\label{equ:cfs_f1}
      F^{1}_{s} = a_{1} \odot \hat{F}^{3d}_{depth} + (1 - a_{1}) \odot \hat{F}^{3d}_{height},     
\end{equation}
where $\odot$ denotes an element-wise product. The first-stage selection operates on the global spatial manifold, which squeezes each feature map of size $H {\times}W{\times}D$ into a scalar.
{The fusion affinity $a_1$, ranging from $0$ to $1$, and its complement, $1{-}a_1$, facilitate a soft selection between $\hat{F}^{3d}_{depth}$ and $\hat{F}^{3d}_{height}$ using weighted averaging.}
This operation emphasizes the large objects like cars and trucks that are distributed globally.
However, the detection of small objects like pedestrians and cyclists is vital for safety-critical intelligent transportation systems.
Therefore, we introduce the second-stage feature selection to aggregate fine-grained local features, {ensuring smaller instances are accurately detected.}
Concretely, in the second stage of CFS, we perform channel pooling and activation on the fused features $F^{1}_{s}$ to obtain the affinity $a_{2}$ in the three-dimensional space and complete the second feature selection:
\begin{equation}
\label{equ:cfs_a2}
      a_{2} = \sigma(\mathcal{G}(\mathcal{P}_{c}(F^{1}_{s}))),    
\end{equation}
where $\mathcal{P}_{c}(\cdot)$ represents the pooling layer across channel dimension, $\mathcal{G}(\cdot)$ denotes the 3D convolution layer with kernel size $7{\times}7{\times}7$. The second-stage selection feature $F^{2}_{s}$ can be obtained by:
\begin{equation}
\label{equ:cfs_f2}
      F^{2}_{s} = a_{2} \odot \hat{F}^{3d}_{depth} + (1 - a_{2}) \odot \hat{F}^{3d}_{height},     
\end{equation}

\begin{figure*}[!t]
\centering
\includegraphics[width=0.8\linewidth]{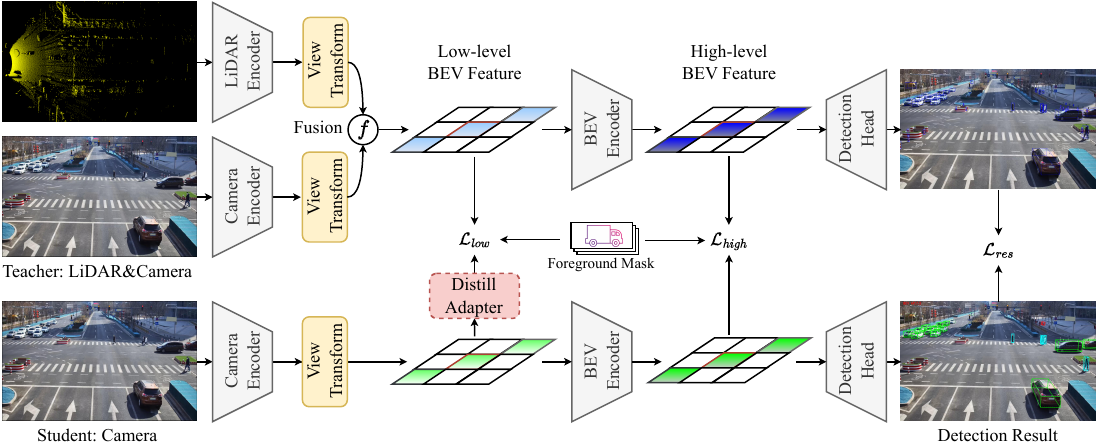}
\vskip -1ex
\caption{\textit{Illustration of the BEV Feature Distillation.}
It employs a fusion-to-camera paradigm, aligning the student CoBEV detector with the LiDAR-camera fusion version of the CoBEV teacher across three stages: low-level BEV feature, high-level BEV feature, and the response. Different from previous work~\cite{zhou2023unidistill}, we only apply supervision signal adaption at low-level features to emphasize valuable street-view structural knowledge of all categories at high-level features and therefore achieve performance improvement agnostic to the object size.
}
\label{overview_distillation}
\vskip -1ex
\end{figure*}

The second-stage feature selection operates on every voxel, thus integrating fine-grained local features adaptively.
Following this, the results of the two feature selections are added through skip connections. To construct the final complementary BEV features $F^{bev}_{com}$, a stride 3D convolution layer is applied to fuse them and reduce the dimension to a $2$-dimensional space:
\begin{equation}
\label{equ:cfs_f_bev}
      F^{bev}_{com} = \mathcal{G}_{fuse}(F^{1}_{s} + F^{2}_{s}), 
\end{equation}
where $\mathcal{G}_{fuse}(\cdot)$ denotes the output 3D convolution layer with stride on the height axis.
Different from the first-stage selection, the second-stage selection focuses on `where' is an informative voxel for the heterogeneous 3D features, therefore complementary to the first stage.

\subsection{BEV Feature Distillation}
\label{sec:bev_feature_distillation}

To fully unleash the potential of the CoBEV model, we introduce the BEV Feature Distillation Framework to further explore the boundaries of monocular camera 3D roadside detection. The framework is demonstrated in Fig.~\ref{overview_distillation}.
Leveraging the unified BEV feature representation, CoBEV seamlessly implements the fusion-to-camera knowledge distillation paradigm, transitioning from the LiDAR-Camera Fusion modality to the Camera modality.
Specifically, the fusion model, serving as the teacher, takes point clouds and image frames as input to provide rigorous supervision to the student CoBEV, across three levels: the low-level BEV features, the high-level BEV features, and the response. The key difference between the teacher and the student model lies in the sparse LiDAR encoder~\cite{yan2018second} and the fusion layer for point clouds and image features, which simply consists of two layers of 2D convolution.
{To ensure effective alignment, we aim at maximizing structural consistency between the teacher and student models.}

Low-level BEV features encapsulate valuable semantic knowledge.
Therefore, during feature distillation at this stage, our focus primarily centers on imitating features of large objects, rather than small objects.
Differing from~\cite{zhou2023unidistill}, we adopt a straightforward MSE loss based on empirical experimentation to guide this process.
Considering the inherent data disparities between sparse LiDAR and dense camera captures, we limit the supervision to task-relevant foreground regions to avoid mimicking irrelevant 3D features in background areas.
To achieve this, we employ the bounding box as the center to generate a Gaussian Mask $M$, assigning weights to the loss in accordance with~\cite{chong2022monodistill}.
In light of occasional instances where the performance of the teacher model may lag behind that of the student model, we introduce an additional Distill Adapter during training to fine-tune the supervision signals.
Importantly, this adapter can be discarded during testing, thus incurring no additional inference resource overhead.
The BEV distillation loss at this stage can be expressed as:
\begin{equation}
\label{equ:loss_low}
      \mathcal{L}_{low}=||M \cdot (F_{teacher}^{low} - \mathcal{G}_{adapt}(F_{student}^{low}) )||_{2}, 
\end{equation}
where $\mathcal{G}_{adapt}(\cdot)$ denotes the distill adapter that consists of three 2D convolutional layers with ReLU activation. The BEV convolutional encoder is realized through three stacked 2D residual blocks. The resultant high-level BEV features effectively encode the structural knowledge of the outdoor scene. Distillation at this stage grants the network an expanded capability to facilitate cross-modality alignment, while concurrently enabling further mimic of all categories of information irrelevant to the object size.
In contrast to the prior work~\cite{zhou2023unidistill}, we refrain from introducing an adapter at this stage because our objective is to maximize the alignment and refinement of all target features, especially those associated with smaller targets. The high-level BEV distillation loss can be expressed as follows:
\begin{equation}
\label{equ:loss_high}
      \mathcal{L}_{high}=||M \cdot (F_{teacher}^{high} - F_{student}^{high} )||_{2}. 
\end{equation}
To align the final output of the student closely with that of the fusion detector, we incorporate additional supervision for both bounding box regression and classification at the response level. In this context, soft labels are employed instead of hard labels, as soft labels have the capacity to convey more informative nuances. Furthermore, soft labels leverage the high confidence associated with definitive outcomes in the teacher model, while attributing lower confidence to uncertain results, effectively serving as a natural filtering mechanism. Following~\cite{hong2022cross}, the response loss can be further dissected into two components: regression loss and classification loss:
\begin{equation}
\label{equ:loss_response}
\begin{aligned}
\mathcal{L}_{res} &= \mathcal{L}_{reg} + \mathcal{L}_{cls} \\ 
                  &= s \times (Smooth L1(b_{t}, b_{s}) + QFL(c_{t}, c_{s})), 
\end{aligned}
\end{equation}
where $s$ represents the IoU confidence score of the soft label bounding box given by the teacher. $Smooth L1(\cdot)$ is the Smooth L1 loss. $b_{t}$ and $b_{s}$ are the bounding box parameters of the soft label and prediction.
$c_{t}$ and $c_{s}$ denote the classification parameters of the teacher and student. $QFL(\cdot)$ represents quality focal loss~\cite{li2020generalized}.

\begin{table*}[ht]
\renewcommand{\arraystretch}{1.3}
    \begin{center}
        \caption{{\textbf{Comparison with the state-of-the-art on the DAIR-V2X-I validation set}~\cite{yu2022dair}\textbf{.}}}
        \label{tab:dair}
        \resizebox{0.9\textwidth}{!}{
\setlength{\tabcolsep}{2mm}{ 
\begin{tabular}{lcc|ccc|ccc|ccc}
\hline
\multirow{2}{*}{\textbf{Method}} & \multirow{2}{*}{\textbf{Modality}} & \multirow{2}{*}{\textbf{Venue}} & \multicolumn{3}{c|}{\textbf{Vehicle $(IoU=0.5)$}} & \multicolumn{3}{c|}{\textbf{Pedestrian $(IoU=0.25)$}} & \multicolumn{3}{c}{\textbf{Cyclist $(IoU=0.25)$}} \\
& & & Easy & Middle & Hard & Easy & Middle & Hard & Easy & Middle & Hard \\
\hline
\hline

PointPillars~\cite{lang2019pointpillars} & LiDAR & CVPR' 19 & 63.07 & 54.00 & 54.01 & 38.53 & 37.20 & 37.28 & 38.46 & 22.60 & 22.49 \\
SECOND~\cite{yan2018second} & LiDAR & Sensors & 71.47 & 53.99 & 54.00 & 55.16 & 52.49 & 52.52 & 54.68 & 31.05 & 31.19 \\
MVXNet~\cite{sindagi2019mvx} & LiDAR \& Camera & ICRA' 19 & 71.04 & 53.71 & 53.76 & 55.83 & 54.45 & 54.40 & 54.05 & 30.79 & 31.06 \\

\hline

ImVoxelNet~\cite{rukhovich2022imvoxelnet} & Camera & WACV' 22 & 44.78 & 37.58 & 37.55 & 6.81 & 6.75 & 6.74 & 21.06 & 13.57 & 13.17 \\
BEVFormer~\cite{li2022bevformer} & Camera & ECCV' 22 & 61.37 & 50.73 & 50.73 & 16.89 & 15.82 & 15.95 & 22.16 & 22.13 & 22.06 \\
BEVDepth~\cite{li2022bevdepth} & Camera & AAAI' 23 & 75.50 & 63.58 & 63.67 & 34.95 & 33.42 & 33.27 & 55.67 & 55.47 & 55.34 \\
BEVHeight~\cite{yang2023bevheight} & Camera & CVPR' 23 & 77.78 & 65.77 & 65.85 & 41.22 & 39.29 & 39.46 & 60.23 & 60.08 & 60.54 \\
BEVHeight++~\cite{yang2023bevheight++} & Camera & - & 79.31 & 68.62 & 68.68 & 42.87 & 40.88 & 41.06 & 60.76 & 60.52 & 61.01 \\

\rowcolor{gray!20}
CoBEV (Ours) & Camera & - & 81.20 & 68.86 & 68.99 & 44.23 & 42.31 & 42.55 & 61.28 & 61.00 & 61.61 \\ %
\rowcolor{gray!20}
CoBEV$_{full}$ (Ours) & Camera & - & \underline{82.01} & \underline{69.57} & \underline{69.66} & \textbf{49.32} & \textbf{47.21} & \textbf{47.48} & \textbf{66.13} & \textbf{66.17} & \textbf{66.69} \\
\rowcolor{gray!20}
\textit{w.r.t. BEVHeight} &  &  & \red{+4.23} & \red{+3.80} & \red{+3.81} & \red{+8.10}& \red{+7.92}& \red{+8.02}& \red{+5.90}& \red{+6.09} & \red{+6.15} \\
\midrule
\rowcolor{gray!20} 
CoBEV* (Ours) & Camera & - & \textbf{82.52} & \textbf{81.48} & \textbf{81.60} & \underline{44.72} & \underline{42.88} & \underline{43.07} & 60.51 & \underline{62.54} & \underline{63.31} \\
\hline

\end{tabular}
}
}
\begin{flushleft}
-- Note: ${full}$ denotes %
using our BEV distillation. * denotes additionally covering the longer range between $100 {\sim} 200m$, while others only cover $0 {\sim} 100m$.
\end{flushleft}

    \end{center}
    \vskip -6ex
\end{table*}

\section{Experiments}
\subsection{Datasets}

\noindent \textbf{DAIR-V2X-I.}
DAIR-V2X~\cite{yu2022dair} is a real-world vehicle-infrastructure collaborative dataset, offering a multi-modal object detection resource within the context of intersection scenes. We focus on the roadside subset, denoted as DAIR-V2X-I. Comprising $10k$ images and corresponding LiDAR point clouds, this subset encompasses a total of $493k$ 3D box annotations, spanning distances from $0$ to $200$ meters, distributed across ten distinct categories. Following previous work~\cite{yang2023bevheight}, we partition DAIR-V2X-I into a training set ($50\%$) and a validation set ($20\%$) to facilitate comparative analyses. Additionally, it is worth noting that the testing examples ($30\%$) are not yet publicly disclosed.

\noindent \textbf{Rope3D.}
Rope3D~\cite{ye2022rope3d} is a large-scale multi-modal 3D detection dataset capturing roadside scenes, comprising a collection of $50k$ camera and LiDAR captures. This dataset offers annotations for $12$ distinct categories within the $0{-}200m$ range, amounting to a substantial total of $1.5M$ 3D box annotations.
Rope3D contains a diverse array of complex traffic scenes, including $26$ different intersection scenarios, spanning conditions such as rainy days, nights, and dawn scenes.
In accordance with the original partitioning strategy detailed in the Rope3D paper~\cite{ye2022rope3d}, we adopt a training-test split that allocates $70\%$ of the images for training and reserves the remaining $30\%$ for testing. Notably, as the LiDAR captures are not publicly accessible, we have refrained from conducting distillation experiments on this dataset.

\noindent \textbf{Supremind-Road.}
The Supremind-Road dataset is a real-world roadside 3D detection dataset that captures authentic traffic scenes. This dataset comprises $16,210$ image frames and corresponding LiDAR scans, delivering annotations for $243k$ 3D object boxes across four classes: \emph{vehicles}, \emph{pedestrians}, \emph{cyclists}, and \emph{tricycles}. Supremind-Road also contains diverse road scenes across different camera parameters, including $130$ different intersection scenarios. Our utilization of this dataset involves testing within genuine application scenarios, thereby validating the adaptability and performance of various algorithms in intelligent transportation systems.
The training, validation, and test data undergo a certain division, following a ratio of $75\%{:}10\%{:}15\%$.
In this setup, the training and validation come from $110$ distinct scenarios, whereas the test set comprises the remaining $20$ different intersections. 
Consequently, the training and validation set form the \texttt{homologous} setting. Conversely, the training and test set constitute the \texttt{heterologous} setting. 
We kindly note that this dataset is not intended for public distribution and does not constitute a contribution to this article.

\begin{table}[t]
\renewcommand{\arraystretch}{1.3}
    \begin{center}
        \caption{{\textbf{Comparison with state-of-the-art methods on the Rope3D validation set}~\cite{ye2022rope3d} \textbf{in \texttt{homologous} settings. +$(G)$ denotes adapting the ground plane.}}}
        \label{tab:rope3d_homo}
        \resizebox{0.50\textwidth}{!}{
\setlength{\tabcolsep}{1mm}{ 
\begin{tabular}{l|cc:cc|cc:cc}
\hline
\multirow{3}{*}{\textbf{Method}} & \multicolumn{4}{c|}{\textbf{$IoU=0.5$}} & \multicolumn{4}{c}{\textbf{$IoU=0.7$}} \\
\cline{2-9}
& \multicolumn{2}{c:}{\textbf{Car}} & \multicolumn{2}{c|}{\textbf{Big Vehicle}} & \multicolumn{2}{c:}{\textbf{Car}} & \multicolumn{2}{c}{\textbf{Big Vehicle}} \\
& AP & Rope & AP & Rope & AP & Rope & AP & Rope \\
\hline
\hline

M3D-RPN+($G$)~\cite{brazil2019m3d} & 54.19 & 62.65 & 33.05 & 44.94 & 16.75 & 32.90 & 6.86 & 24.19 \\
Kinematic3D+($G$)~\cite{brazil2020kinematic} & 50.57 & 58.86 & 37.60 & 48.08 & 17.74 & 32.90 & 6.10 & 22.88 \\
MonoDLE+($G$)~\cite{ma2021delving} & 51.70 & 60.36 & 40.34 & 50.07 & 13.58 & 29.46 & 9.63 & 25.80 \\
BEVFormer~\cite{li2022bevformer} & 50.62 & 58.78 & 34.58 & 45.16 & 24.64 & 38.71 & 10.05 & 25.56 \\
BEVDepth~\cite{li2022bevdepth} & 69.63 & 74.70 & 45.02 & 54.64 & 42.56 & 53.05 & 21.47 & 35.82 \\
BEVHeight~\cite{yang2023bevheight} & \textbf{74.60} & \textbf{78.72} & \underline{48.93} & \underline{57.70} & \underline{45.73} & \underline{55.62} & \underline{23.07} & \underline{37.04} \\

\hline

\rowcolor{gray!20}
CoBEV (Ours) & \underline{73.39} & \underline{77.11} & \textbf{52.77} & \textbf{60.28} & \textbf{52.72} & \textbf{60.57} & \textbf{29.28} & \textbf{41.52} \\    %
\rowcolor{gray!20}
\textit{w.r.t. BEVHeight} & \green{-1.21} & \green{-1.61} & \red{+3.84} & \red{+2.58}& \red{+6.99}& \red{+4.95}& \red{+6.21}& \red{+4.48} \\

\hline

\end{tabular}
}
}

    \end{center}
    \vskip -4ex
\end{table}

\begin{table}[ht]
\renewcommand{\arraystretch}{1.3}
    \begin{center}
        \caption{{\textbf{Comparison with state-of-the-art methods on the Rope3D validation set}~\cite{ye2022rope3d} \textbf{in \texttt{heterologous} settings.}}}
        \label{tab:rope3d_heter}
        \resizebox{0.50\textwidth}{!}{
\setlength{\tabcolsep}{0.5mm}{ 
\begin{tabular}{lc|cc:cc|cc:cc}
\hline
\multirow{3}{*}{\textbf{Method}} & \multirow{3}{*}{\textbf{Lifting}} & \multicolumn{4}{c|}{\textbf{$IoU=0.5$}} & \multicolumn{4}{c}{\textbf{$IoU=0.7$}} \\
\cline{3-10}
& & \multicolumn{2}{c:}{\textbf{Car}} & \multicolumn{2}{c|}{\textbf{Big Vehicle}} & \multicolumn{2}{c:}{\textbf{Car}} & \multicolumn{2}{c}{\textbf{Big Vehicle}} \\
& & AP & Rope & AP & Rope & AP & Rope & AP & Rope \\
\hline
\hline

BEVFormer~\cite{li2022bevformer} & Implicit & 25.98 & 39.51 & 8.81 & 24.67 & 3.87 & 21.84 & 0.84 & 18.42 \\
BEVDepth~\cite{li2022bevdepth} & Explicit & 9.00 & 25.80 & 3.59 & 20.39 & 0.85 & 19.38 & 0.30 & 17.84 \\
BEVHeight~\cite{yang2023bevheight} & Explicit & 29.65 & 42.48 & 13.13 & 28.08 & 5.41 & 23.09 & 1.16 & 18.53 \\

\hline

\rowcolor{gray!20}
CoBEV (Ours) & Hybrid & \textbf{31.25} & \textbf{43.74} & \textbf{16.11} & \textbf{30.73} & \textbf{6.59} & \textbf{24.01} & \textbf{2.26} & \textbf{19.71} \\   %
\rowcolor{gray!20}
\textit{w.r.t. BEVHeight} & - & \red{+1.60} & \red{+1.26} & \red{+2.98} & \red{+2.65}& \red{+1.18}& \red{+0.92}& \red{+1.10}& \red{+1.18} \\

\hline

\end{tabular}
}
}

    \end{center}
    \vskip -6ex
\end{table}

\subsection{Comparison with the State-of-the-Arts}
For a comprehensive evaluation, we compare the proposed CoBEV against state-of-the-art BEV detectors, including BEVFormer~\cite{li2022bevformer}, BEVDepth~\cite{li2022bevdepth}, and BEVHeight~\cite{yang2023bevheight}. The evaluation is conducted on three datasets described as follows.

\noindent\textbf{Results on the DAIR-V2X-I dataset.}
In Tab.~\ref{tab:dair}, we report the results of LiDAR-based~\cite{lang2019pointpillars,yan2018second} and multi-modal~\cite{sindagi2019mvx} detectors, reproduced by the original DAIR-V2X-I dataset~\cite{yu2022dair}.
Our results demonstrate that CoBEV outperforms all other methods across the board, with consistent and notable improvements in each category. 
{Specifically, it achieves a performance boost of ${+}3.09\%$ ($68.86\%$ \textit{vs.} $65.77\%$) on the \textit{vehicle} category, ${+}3.02\%$ ($42.31\%$ \textit{vs.} $39.29\%$) on the \textit{pedestrain} category, and ${+}0.92\%$ ($61.00\%$ \textit{vs.} $60.08\%$) on the \textit{cyclist} category, as compared to the previously best published detector~\cite{yang2023bevheight}. Additionally, we have included the results of BEVHeight++~\cite{yang2023bevheight++} in our experimental tables. As shown, even as \textit{concurrent} works, CoBEV demonstrates significant improvements compared to BEVHeight++, outperforming it by $2.7\% / 6.45\% / 5.37\%$ on AP for Vehicle, Pedestrian, and Cyclist, respectively.}
Remarkably, CoBEV marks the first approach where a camera-based monocular 3D detector achieves \textit{vehicle} detection accuracy exceeding $80\%$ in the Easy setting.
Moreover, 
the accuracy of CoBEV for \textit{vehicles} and \textit{cyclists} largely surpasses that of LiDAR-based and multi-modal methods by ${\sim}10\%$. 
When our BEV feature distillation framework is fully equipped, the detection accuracy of CoBEV$_{full}$ has seen substantial enhancements across all categories, especially on small targets, which is comparable to the LiDAR detectors ($47.21\%$ \textit{vs.} $52.49\%$).
This improvement is attributed to the richer knowledge possessed by the fusion-modal teacher when it comes to detecting small targets.
Notably, CoBEV$_{full}$ establishes new state-of-the-art performance on the DAIR-V2X-I benchmark.
We notice that previous methods were trained only using ground truth labels within the $0{\sim}100m$ range for supervision. This is due to the fact that labels within the $100{\sim}200m$ range mainly contain challenging long-distance \textit{vehicles} and rarely consist of small targets like \textit{pedestrians}.
To address this, we for the first time, cover a longer range of BEV detection tasks, encompassing targets in the $100{\sim}200m$.
The method is denoted as CoBEV* in Tab.~\ref{tab:dair}. 
In comparison to CoBEV, CoBEV* demonstrates improvements across all categories, especially in the \textit{vehicle} category ($81.48\%$ \textit{vs.} $68.86\%$). Obtaining better performance in a longer distance demonstrates the advance of our hybrid lifting method and the effectiveness of fusing complimentary features based on depth and height.

\begin{figure*}[!t]
\centering
\includegraphics[width=0.9\linewidth]{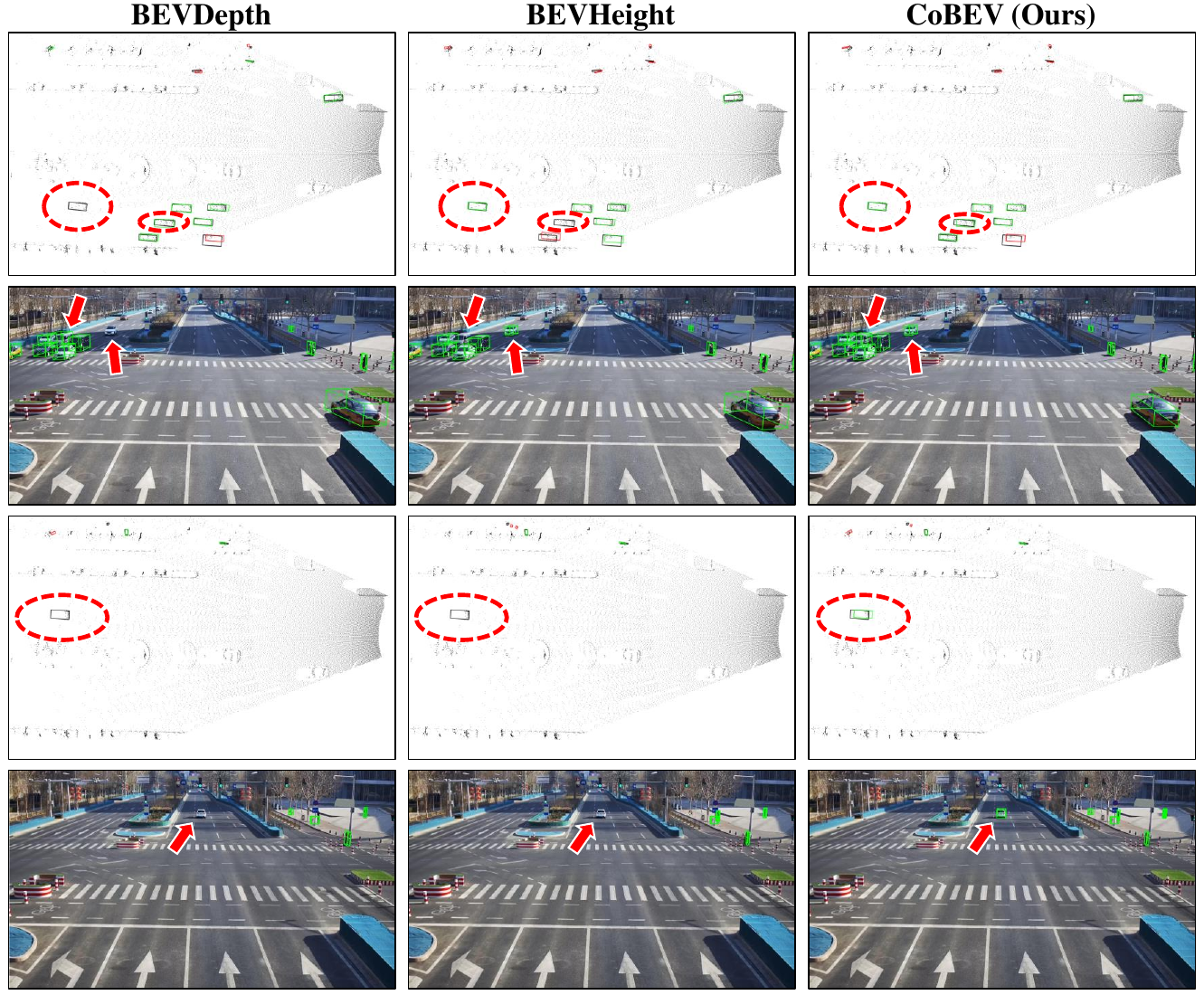}
\vskip -0.5ex
\caption{{\textit{Qualitative comparisons on the DAIR-V2X-I dataset~\cite{yu2022dair}.} BEVDepth~\cite{li2022bevdepth} and BEVHeight~\cite{yang2023bevheight} exhibit missed detections when encountering distant targets. In contrast, CoBEV demonstrates the capability to effectively address challenging long-distance targets, surpassing the performance of prior methods.}
}
\label{figure:dair_comp}
\vskip -1.5ex
\end{figure*}

{Fig.~\ref{figure:dair_comp} shows a comparison of previous methods on the DAIR-V2X-I dataset. In the BEV LiDAR view, the green represents the bounding box detected by the model, and the red represents the ground truth. In the roadside camera view, the 3D box represents the bounding box detected by the model. 
{BEVDepth misses long-range targets, whereas CoBEV accurately detects challenging distant objects, aligning with our quantitative findings.}
}

\noindent\textbf{Results on the Rope3D dataset.} 
On the Rope3D dataset~\cite{ye2022rope3d}, we conduct a comparative evaluation of CoBEV against state-of-the-art monocular 3D detectors, considering two distinct settings: \texttt{homologous} and \texttt{heterologous}.
Notably, some detection methods incorporate the ground plane as an additional constraint, marked as +$(G)$.
As shown in Tab.~\ref{tab:rope3d_homo}, CoBEV achieves the second-best \textit{car} detection accuracy ($73.39\%$ \textit{vs.} $74.60\%$) and the top accuracy for \textit{big vehicles} ($52.77\%$ \textit{vs.} $48.93\%$) under the easy evaluation conditions characterized by $IoU{=}0.5$. 
Under the more strict evaluation of $IoU{=}0.7$, CoBEV outperforms BEVHeight by large margins of $6.99\%$ / $4.95\%$ and $6.21\%$ / $4.48\%$ in terms of $AP_{3D|R40}$ and $Rope_{score}$ for \textit{car} and \textit{big vehicle} categories, respectively.
Moving to the challenging heterologous setting, as displayed in Tab.~\ref{tab:rope3d_heter}, all monocular 3D detectors endure a significant performance drop.
Consistent with the previous comparison, CoBEV maintains its overall superiority, exhibiting state-of-the-art accuracy under both easy and strict evaluation conditions.
{This exceptional performance is due to CoBEV's camera-aware hybrid lifting design and the complementary feature selection module, which together build robust BEV features.}
Consequently, in the heterologous setting with new viewing cameras and unseen intersection scenes, CoBEV shows enhanced generalization capabilities across camera parameters and traffic scenarios.

\noindent\textbf{Results on the Supremind-Road dataset.} 
To facilitate a comprehensive evaluation 
across diverse intersection scenarios, we reproduce the previous state-of-the-art BEV detectors on our new Supremind-Road dataset. As shown in Tab.~\ref{tab:supremind}, CoBEV emerges as the top-performing method across the \textit{vehicle}, \textit{pedestrian}, \textit{cyclist}, and \textit{tricycle} categories, encompassing various difficulty levels within the \texttt{homologous} setting. 
Notably, it outperforms the second-best BEVHeight by considerable margins, with improvements of ${+}1.47\%$ / ${+}1.54\%$, ${+}2.45\%$ / ${+}1.06\%$, ${+}2.97\%$ / ${+}1.00\%$, and ${+}4.19\%$ / ${+}5.15\%$, respectively.
Under the challenging \texttt{heterologous} setting, although the performances of all detectors drop greatly, CoBEV still achieves state-of-the-art performance across most categories. Especially noteworthy is its performance in large targets of the \textit{vehicle} and \textit{tricycle}, where it shows significant advantages of ${+}2.80\%$ / ${+}2.24\%$ and ${+}7.65\%$ / ${+}7.96\%$, respectively. This proves CoBEV's better robustness and generalization capabilities to previously unseen transportation scenarios.
In cases involving smaller targets such as \textit{pedestrians} and \textit{cyclists}, CoBEV's accuracy is on par with BEVHeight, significantly surpassing BEVFormer and BEVDepth.

\begin{table*}[t]
\renewcommand{\arraystretch}{1.3}
    \begin{center}
        \caption{\textbf{Mono3D detection results on the Supremind-Road dataset in \texttt{homologous} and \texttt{heterologous} settings.}}
        \label{tab:supremind}
        \resizebox{0.9\textwidth}{!}{
\setlength{\tabcolsep}{2mm}{    
\begin{tabular}{lc|cccccccc:cccccccc}
\hline
\multirow{3}{*}{\textbf{Method}} & \multirow{3}{*}{\textbf{Lifting}} & \multicolumn{8}{c:}{Validation Set (\texttt{homologous})} & \multicolumn{8}{c}{Test Set (\texttt{heterologous})}  \\

& & \multicolumn{2}{c}{\textbf{Vehicle}} & \multicolumn{2}{c}{\textbf{Pedestrian}} & \multicolumn{2}{c}{\textbf{Cyclist}} & \multicolumn{2}{c:}{\textbf{Tricycle}} & \multicolumn{2}{c}{\textbf{Vehicle}} & \multicolumn{2}{c}{\textbf{Pedestrian}} & \multicolumn{2}{c}{\textbf{Cyclist}} & \multicolumn{2}{c}{\textbf{Tricycle}} \\

& & Easy & Hard & Easy & Hard & Easy & Hard & Easy & Hard & Easy & Hard & Easy & Hard & Easy & Hard & Easy & Hard \\
\hline
\hline

BEVFormer~\cite{li2022bevformer} & Implicit & 69.17 & 58.78 & 11.32 & 11.35 & 25.44 & 20.14 & 20.74 & 19.60 & 46.96 & 37.87 & 0.12 & 0.12 & 2.62 & 2.59 & 0.00 & 0.00 \\
BEVDepth~\cite{li2022bevdepth} & Explicit & 73.74 & 63.21 & 14.71 & 14.25 & 24.73 & 19.52 & 16.89 & 18.58 & 12.95 & 10.22 & 0.06 & 0.06 & 1.04 & 1.00 & 0.83 & 0.83 \\
BEVHeight~\cite{yang2023bevheight} & Explicit & 76.65 & 66.17 & 20.96 & 20.59 & 50.01 & 44.20 & 43.68 & 43.03 & 52.44 & 41.86 & 1.41 & 0.94 & \textbf{18.61} & \textbf{16.74} & 23.07 & 20.42 \\

\hline

\rowcolor{gray!20}
CoBEV (Ours) & Hybrid & \textbf{78.12} & \textbf{67.71} & \textbf{23.41} & \textbf{21.65} & \textbf{52.98} & \textbf{45.20} & \textbf{47.87} & \textbf{48.18} & \textbf{55.24} & \textbf{44.10} & \textbf{1.46} & \textbf{1.32} & 18.32 & 16.46 
 & \textbf{30.72} & \textbf{28.38} \\
 \rowcolor{gray!20}
\textit{w.r.t. BEVHeight} & - & \red{+1.47} & \red{+1.54} & \red{+2.45} & \red{+1.06} & \red{+2.97} & \red{+1.00} & \red{+4.19} & \red{+5.15} & \red{+2.80} & \red{+2.24} & \red{+0.05} & \red{+0.38} & \green{-0.29} & \green{-0.28} & \red{+7.65} & \red{+7.96} \\

\hline

\end{tabular}
}
}

    \end{center}
    \vskip -1ex
\end{table*}

\begin{table*}[t]
\renewcommand{\arraystretch}{1.3}
    \begin{center}
        \caption{\textbf{Robustness analysis on the DAIR-V2X-I validation set.}
        Three disturbed factors of roadside cameras are investigated, including focal length, roll angle, and pitch angle.}
        \label{tab:robust}
        \resizebox{0.9\textwidth}{!}{
\setlength{\tabcolsep}{2mm}{ 
\begin{tabular}{lccc|ccc|ccc|ccc}
\hline
\multirow{2}{*}{\textbf{Method}} & \multicolumn{3}{c|}{\textbf{Disturbed}} & \multicolumn{3}{c|}{\textbf{Vehicle $(IoU=0.5)$}} & \multicolumn{3}{c|}{\textbf{Pedestrian $(IoU=0.25)$}} & \multicolumn{3}{c}{\textbf{Cyclist $(IoU=0.25)$}} \\
& focal & roll & pitch & Easy & Middle & Hard & Easy & Middle & Hard & Easy & Middle & Hard \\
\hline
\hline

\multirow{6}{*}{BEVDepth~\cite{li2022bevdepth}} & - & - & - & 75.31 & 65.24 & 65.32 & 32.68 & 31.01 & 31.33 & 46.96 & 50.88 & 51.44 \\
 & \checkmark & - & - & 72.17 & 60.19 & 60.20 & 25.75 & 25.16 & 24.35 & 40.65 & 47.09 & 47.21 \\
 & - & \checkmark & - & 74.78 & 62.72 & 62.81 & 30.80 & 30.20 & 30.43 & 45.58 & 50.07 & 50.72 \\
 & - & - & \checkmark & 74.83 & 62.76 & 62.85 & 30.21 & 28.62 & 28.91 & 46.07 & 50.15 & 50.85 \\
 & - & \checkmark & \checkmark & 74.62 & 62.57 & 62.66 & 30.38 & 28.87 & 29.13 & 45.96 & 50.15 & 50.79 \\
 \rowcolor{gray!20}
 & \checkmark & \checkmark & \checkmark & 71.91 & 59.94 & 59.96 & 26.61 & 25.18 & 25.22 & 39.79 & 46.11 & 46.13 \\

\hline

\multirow{6}{*}{BEVHeight~\cite{yang2023bevheight}} & - & - & - & 78.08 & 65.97 & 66.04 & 40.01 & 38.21 & 38.38 & 58.01 & 60.46 & 60.95 \\
 & \checkmark & - & - & 72.30 & 60.45 & 60.47 & 32.18 & 30.65 & 29.65 & 50.06 & 55.04 & 55.14 \\
 & - & \checkmark & - & 77.65 & 65.57 & 65.65 & 38.38 & 36.60 & 36.72 & 56.15 & 59.11 & 59.52 \\
 & - & - & \checkmark & 75.37 & 63.31 & 63.38 & 33.13 & 31.47 & 31.63 & 52.88 & 56.07 & 56.44 \\
 & - & \checkmark & \checkmark & 75.06 & 63.08 & 63.16 & 33.67 & 31.19 & 31.30 & 51.65 & 54.93 & 56.83 \\
 \rowcolor{gray!20}
 & \checkmark & \checkmark & \checkmark & 71.71 & 59.92 & 59.96 & 27.81 & 26.43 & 26.36 & 47.42 & 51.19 & 51.26 \\

\hline

\multirow{7}{*}{CoBEV (Ours)} & - & - & - & 81.20 & 68.86 & 68.99 & 44.23 & 42.31 & 42.55 & 61.28 & 61.00 & 61.61 \\
 & \checkmark & - & - & 78.70 & 66.36 & 66.43 & 36.19 & 34.36 & 34.39 & 55.56 & 57.11 & 57.39 \\
 & - & \checkmark & - & 81.03 & 68.78 & 68.91 & 42.47 & 40.56 & 40.88 & 61.38 & 61.94 & 62.59 \\
 & - & - & \checkmark & 78.57 & 66.33 & 66.45 & 36.82 & 35.01 & 35.51 & 57.65 & 58.59 & 59.28 \\
 & - & \checkmark & \checkmark & 78.60 & 66.34 & 66.47 & 37.38 & 35.57 & 35.89 & 56.44 & 57.44 & 58.10 \\
 \rowcolor{gray!20}
 & \checkmark & \checkmark & \checkmark & \textbf{75.53} & \textbf{63.46} & \textbf{63.55} & \textbf{30.75} & \textbf{30.08} & \textbf{29.17} & \textbf{51.42} & \textbf{54.78} & \textbf{54.97} \\
 \rowcolor{gray!20}
 & \multicolumn{3}{c|}{\textit{w.r.t. BEVHeight}} & \red{+3.82} & \red{+3.54} & \red{+3.59} & \red{+2.94} & \red{+3.65} & \red{+2.81} & \red{+4.00} & \red{+3.59} & \red{+3.71} \\
\hline

\end{tabular}
}
}

    \end{center}
    \vskip -4ex
\end{table*}

\noindent {
\noindent\textbf{Timing and Parameter Counts.} The number of parameters and computational latency of CoBEV were evaluated using an RTX 4090 GPU. Compared to BEVHeight, CoBEV's parameter count increases primarily in the view transformation (${+}24.7M$) and feature fusion (${+}14.1M$) modules. With ResNet50 as the backbone, CoBEV contains $115M$ parameters. Given an input image resolution of $864{\times}1536$ with a batch size of $1$, and a BEV grid resolution of $0.8m \times 0.8m$, without further model quantization and pruning, CoBEV achieves an image throughput of $27.78$ FPS under mixed precision. Using ResNet101 as the backbone, CoBEV contains $134M$ parameters and achieves an image throughput of $25.64$ FPS. 
Both configurations meet the real-time performance requirements for roadside deployment.
}

\subsection{Robustness Analysis}
In real-world transportation scenarios, cameras positioned at intersections are often subjected to variations in extrinsic parameters caused by factors such as wind, vibrations, human adjustments, and other environmental conditions. 
Additionally, the intrinsic parameters of different cameras will also change. 
This motivates us to investigate the robustness of different BEV detectors in the context of fluctuations in camera parameters.
Following the simulation methodology outlined in~\cite{yu2023benchmarking}, we introduce offset noise with a $N(0, 1.67)$ distribution to the \textit{roll} and \textit{pitch} angles associated with the extrinsic matrix. 
For the camera \textit{focal} length, we introduce scale noise, with the scaling coefficient following a $N(1, 0.2)$ distribution.
Corresponding rotation and scaling transformations are also applied to the camera capture to ensure that the image is accurately mapped to the reference coordinate system using the camera parameters, thus preserving the calibration relationship.

\begin{figure}[!t]
\centering
\includegraphics[width=0.9\linewidth]{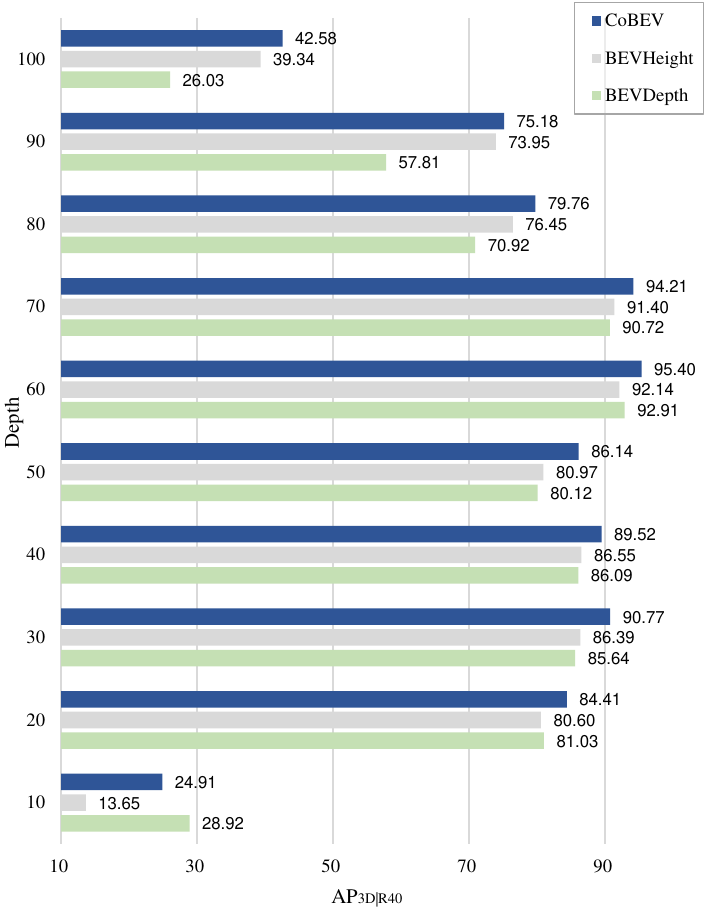}
\vskip -1ex
\caption{\textit{Range-wise evaluation on the DAIR-V2X-I validation set.} Metric is $AP_{3D|R40}$ of the Vehicle category under moderate setting. The sample interval is $10m$, \textit{e.g.}, the value at vertical axis $50$ indicates the overall performance of all samples between $45m$ and $55m$.
}
\vskip -3ex
\label{fig:depth-ap}
\end{figure}

To ensure a fair and consistent comparison, BEVDepth~\cite{li2022bevdepth}, BEVHeight~\cite{yang2023bevheight}, and the proposed CoBEV all adapt exactly the same camera parameter perturbation during the training process.
As detailed in Tab.~\ref{tab:robust}, CoBEV maintains the best accuracy across all test-time scenarios involving noisy camera parameters.
When only the camera focal length is disturbed, CoBEV exhibits significantly enhanced robustness compared to BEVHeight and BEVDepth, with respective accuracies of $66.36\%$ \textit{vs.} $60.45\%$ \textit{vs.} $60.19\%$.
Similarly, when noise perturbs solely the camera angles, CoBEV maintains its superiority in terms of accuracy ($66.33\%$ \textit{vs.} $63.08\%$ \textit{vs.} $62.57\%$).
Furthermore, in situations where all camera parameter noise is considered, CoBEV clearly stands out as the only method achieving Vehicle and Pedestrian detection accuracy exceeding $60\%$ ($63.46\%$ \textit{vs.} $59.92\%$ \textit{vs.} $59.94\%$) and $30\%$ ($30.08\%$ \textit{vs.} $26.43\%$ \textit{vs.} $25.18\%$), respectively.
These results reveal CoBEV's excellent robustness and resistance to interference when confronted with varying camera parameters.

{Additionally, we qualitatively visualize CoBEV's detection robustness under inaccurate camera calibrations. As shown in Fig.~\ref{figure:robust_comp}, we evaluate the robustness of 3D detection under changes in camera parameters such as roll, pitch, and focal length, which may vary due to calibration errors or external factors like wind-induced vibrations. These parameter changes make accurate 3D feature extraction more challenging, leading to decreased detection accuracy for distant vehicles in BEVDepth and BEVHeight. When the camera focal length increases, the field of view decreases, causing close-range targets to be truncated. CoBEV successfully detects these truncated targets, demonstrating its robustness to camera calibration errors.}

\noindent {
\noindent{\textbf{Handling dynamic weathers and lighting conditions.}} 
CoBEV is evaluated on three real-world datasets that inherently contain dynamically changing lighting conditions (day, dusk, night) and weather (sunny, foggy, rainy). Although recent advancements in 3D detection have shown significant improvements on classic vehicle-centric datasets such as KITTI~\cite{geiger2012we}, nuScene~\cite{caesar2020nuscenes}, and Waymo~\cite{sun2020scalability}, the generalization of existing data-driven methods under extreme weather conditions remains unsatisfactory~\cite{kilic2021lidar,hahner2022lidar,hahner2021fog}. Consequently, some methods have developed datasets specifically designed for vehicle detection in complex weather environments~\cite{hendrycks2019scaling,dong2023benchmarking}. The roadside datasets discussed in this paper, collected under a wide variety of weather and lighting conditions without meticulous weather classification per frame, preclude quantitative evaluation, necessitating our use of qualitative comparisons to demonstrate CoBEV's robustness.
As shown in Fig.~\ref{figure:weather_comp}, we present detection results for \textit{night}, \textit{dusk}, \textit{rainy}, and \textit{foggy} scenes, comparing BEVDepth~\cite{li2022bevdepth} and BEVHeight~\cite{yang2023bevheight}. In these challenging scenarios, CoBEV correctly detects partially occluded cars, as well as difficult small targets (pedestrians) and vehicles at long distances, due to its robust BEV features. This demonstrates the value of CoBEV in real-world applications under different weather and lighting conditions.
}

\begin{figure}[!t]
\centering
\includegraphics[width=0.9\linewidth]{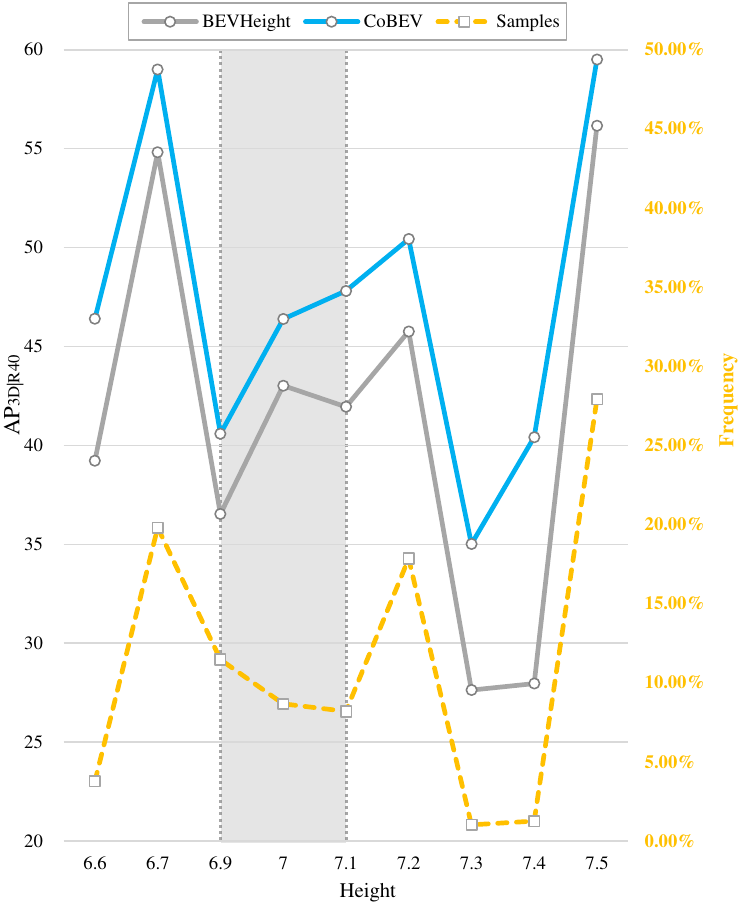}
\vskip -1.5ex
\caption{\textit{Height-wise evaluation on the Rope3D validation set.} Metric is $AP_{3D|R40}$ of all vehicles under $IoU{=}0.7$ setting. The sample interval is $0.1m$, \textit{e.g.}, the value at horizontal axis $7.0$ indicates the overall performance between $6.95m$ and $7.05m$. The overall accuracy is associated with the training set's sample frequency. However, we observed a negative correlation within the height range of $6.9{\sim}7.1$ meters, suggesting that the detection error based on height increases as the camera height decreases.
}

\vskip -3ex
\label{fig:height-ap}
\end{figure}

\begin{figure*}[!t]
\centering
\includegraphics[width=0.85\linewidth]{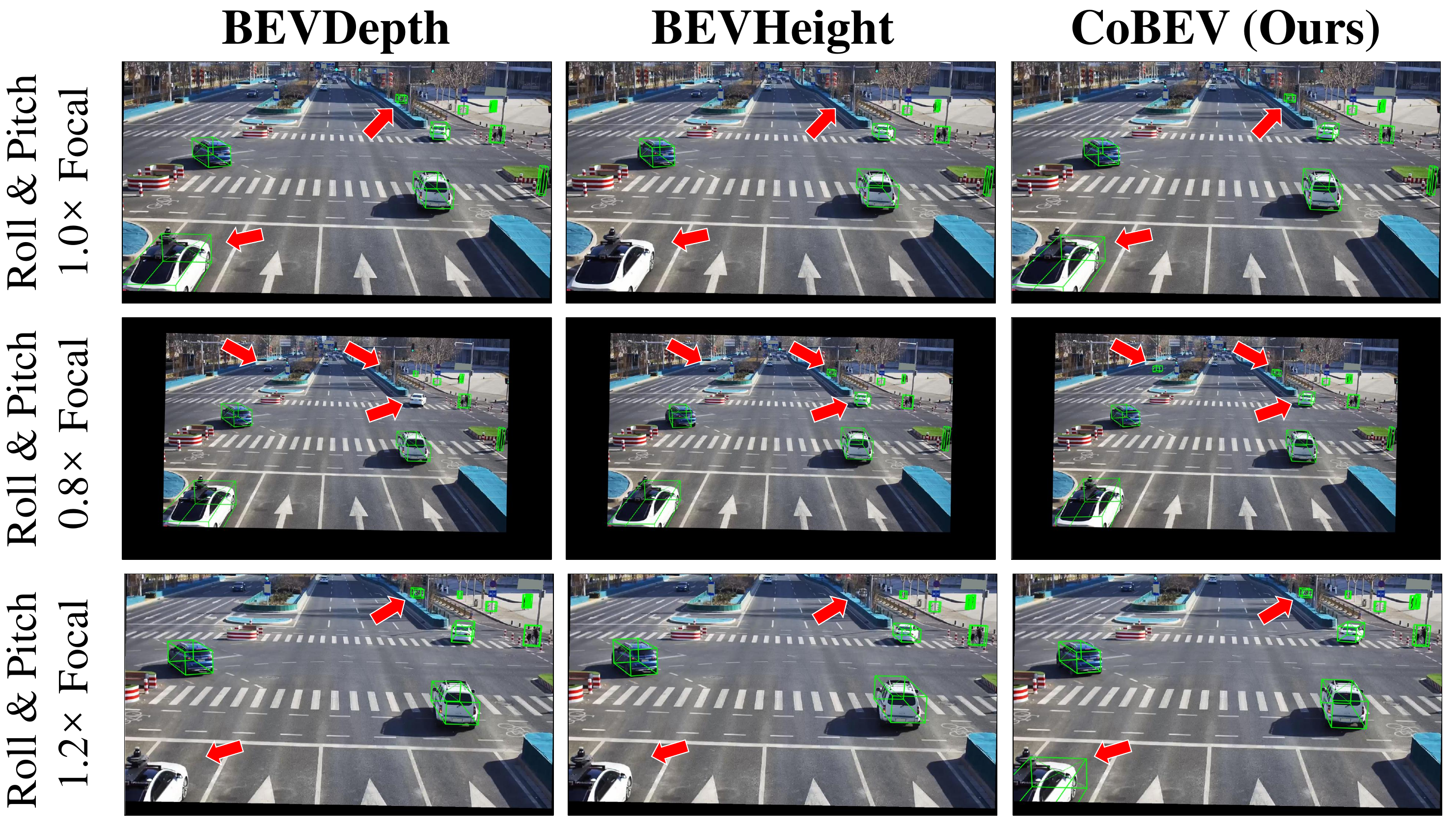}
\vskip -2ex
\caption{{\textit{Qualitative comparisons on the DAIR-V2X-I dataset under the camera parameters disturbance.} In the condition of various camera parameter disturbances, including focal length, roll, and pitch, CoBEV consistently demonstrates the most robust target detection capabilities. This applies to both targets truncated at close range and small targets situated at long distances.}
}
\label{figure:robust_comp}
\vskip -1ex
\end{figure*}

\noindent{\textbf{Delving into the complementary attribute of depth-based and height-based BEV detector.}} 
To delve deeper into the complementary attribute between depth- and height detectors, we present the accuracy distributions of BEVDepth~\cite{li2022bevdepth}, BEVHeight~\cite{yang2023bevheight}, and the proposed CoBEV within different range intervals.
As shown in Fig.~\ref{fig:depth-ap}, it becomes apparent that depth-based detectors exhibit a notable advantage in short-range scenarios, particularly at distances of around $10$ meters.
We believe that this advantage stems from the fact that many targets within this range may be partially obstructed by the camera's Field-of-View (FoV). 
In contrast, BEVHeight prioritizes capturing the overall height of each target to facilitate classification, emphasizing semantic information.
Thus, height-based detectors are more focused on the semantic aspect of targets. Estimating semantic information for truncated objects is more challenging than regressing depth at short distances, which accounts for the dominance of depth-based detectors in this range. CoBEV, however, attains a harmonious balance in accuracy.

As the target distance exceeds $70$ meters, regressing depth becomes increasingly challenging, affording height-based detectors a significant advantage. 
The robust complementary BEV features constructed by CoBEV further amplify this advantage. 
Fig.~\ref{fig:height-ap} presents the accuracy distribution of BEVHeight and CoBEV within different camera height intervals, along with the frequency of training samples. 
In general, detection accuracy shows a positive correlation with the frequency of training samples. 
However, it is worth noting an abnormal negative correlation between accuracy and frequency in the $6.85{\sim}7.15$ meter interval.
This is attributed to the gradual reduction in camera installation height from $7.15$ meters to $6.85$ meters, leading to increased height-based detection errors that intensify the difficulty of the task, even surpassing the influence of training sample frequency. 
CoBEV leverages depth-based complementary features and it continues to maintain a significant advantage within relatively challenging intervals.

\begin{figure*}[!t]
\centering
\includegraphics[width=0.85\linewidth]{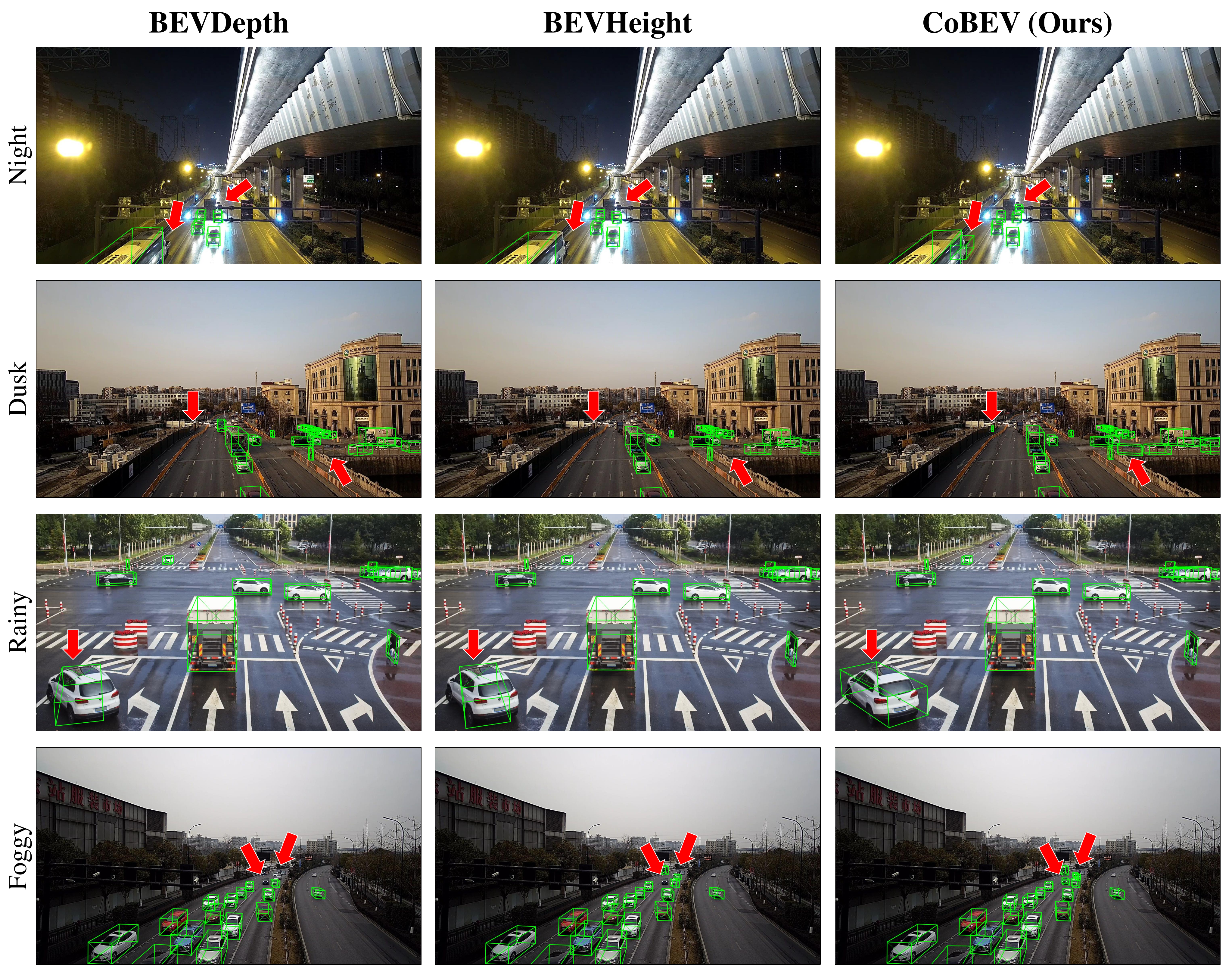}
\vskip -1ex
\caption{{\textit{Qualitative comparisons on dynamic changes in lighting conditions and various weather scenarios.} CoBEV correctly detects partially occluded vehicles as well as distant vehicles and pedestrians in night, dusk, rainy, and foggy scenes.}
}
\label{figure:weather_comp}
\vskip -1ex
\end{figure*}

\begin{table*}[t]
\renewcommand{\arraystretch}{1.3}
    \begin{center}
        \caption{{\textbf{Ablations on BEV feature fusion on the DAIR-V2X-I dataset}~\cite{yu2022dair}\textbf{.}}}
        \label{tab:2d-3d-fusion}
        \resizebox{1.0\textwidth}{!}{
\setlength{\tabcolsep}{2mm}{ 
\begin{tabular}{llcc|ccc|ccc|ccc}
\hline
\multirow{2}{*}{\textbf{Source}} & \multirow{2}{*}{\textbf{Aggregation}} & \multirow{2}{*}{\textbf{Depth-based}} & \multirow{2}{*}{\textbf{Height-based}} & \multicolumn{3}{c|}{\textbf{Vehicle $(IoU=0.5)$}} & \multicolumn{3}{c|}{\textbf{Pedestrian $(IoU=0.25)$}} & \multicolumn{3}{c}{\textbf{Cyclist $(IoU=0.25)$}} \\
& & & & Easy & Middle & Hard & Easy & Middle & Hard & Easy & Middle & Hard \\
\hline
\hline

\multirow{2}{*}{Vanilla} & \multirow{2}{*}{$w/o$} & \checkmark & - & 75.50 & 63.58 & 63.67 & 34.95 & 33.42 & 33.27 & 55.67 & 55.47 & 55.34 \\   %
  & & - & \checkmark & 77.78 & 65.77 & 65.85 & 41.22 & 39.29 & 39.46 & 60.23 & 60.08 & 60.54 \\  %

\hline
\multirow{2}{*}{2D BEV} & CBAM~\cite{woo2018cbam} & \checkmark & \checkmark & 78.25 & 66.15 & 66.23 & 41.45 & 39.57 & 39.81 & 56.92 & 58.05 & 58.52 \\
& Attention Feature Fusion~\cite{dai2021attentional} & \checkmark & \checkmark & 78.61 & 66.41 & 66.49 & 42.06 & 40.16 & 40.36 & 59.35 & 60.15 & 60.72 \\  %

\hline
\multirow{6}{*}{3D Voxel} & Add & \checkmark & \checkmark & 78.55 & 68.53 & 68.65 & \textbf{45.38} & \textbf{43.33} & \textbf{43.66} & 60.25 & 58.65 & 59.26 \\
& Concatenation & \checkmark & \checkmark & 78.53 & 68.46 & 68.59 & 44.55 & 42.48 & 42.79 & 60.87 & 60.22 & 60.92 \\
& Channel Attention$_{3D}$ & \checkmark & \checkmark & 80.87 & 68.59 & 68.71 & 44.04 & 42.00 & 42.35 & 60.85 & 58.08 & 60.06 \\
& Spatial Attention$_{3D}$ & \checkmark & \checkmark & 81.00 & 68.74 & 68.86 & 43.56 & 41.63 & 41.91 & 60.68 & 60.36 & 61.13 \\ %
& \cellcolor{gray!20}CFS (Ours) & \cellcolor{gray!20}\checkmark & \cellcolor{gray!20}\checkmark & \cellcolor{gray!20}\textbf{81.20} & \cellcolor{gray!20}\textbf{68.86} & \cellcolor{gray!20}\textbf{68.99} & \cellcolor{gray!20}44.23 & \cellcolor{gray!20}42.31 & \cellcolor{gray!20}42.55 & \cellcolor{gray!20}\textbf{61.28} & \cellcolor{gray!20}\textbf{61.00} & \cellcolor{gray!20}\textbf{61.61} \\
& \textit{w.r.t. BEVHeight} & - & - & \red{+3.42} & \red{+3.09} & \red{+3.14} & \red{+3.01}& \red{+3.02}& \red{+3.09}& \red{+1.05}& \red{+0.92} & \red{+1.07} \\

\hline

\end{tabular}
}
}
    \end{center}
    \vskip -6ex
\end{table*}

\subsection{Ablation Studies}
We conduct a series of ablations to verify the core components of CoBEV and present the findings of each experiment:

\noindent\textbf{Fusion Strategy.} 
We explore different fusion strategies for depth-based and height-based features using the ResNet101 image backbone~\cite{he2016deep}. Tab.~\ref{tab:2d-3d-fusion} reveals that incorporating 3D features obtained via our partial-pillar voxel pooling for fusion outperforms the approach of obtaining 2D features directly through voxel pooling and then fusing. 
For instance, merely adding 3D features leads to significant enhancements in Vehicle ($68.53\%$ \textit{vs.} $65.77\%$) and Pedestrian ($43.33\%$ \textit{vs.} $39.29\%$) detection accuracy. 
The proposed Complementary Feature Selection (CFS) module achieves the best Vehicle ($68.86\%$ \textit{vs.} $65.77\%$) and Cyclist ($61.00\%$ \textit{vs.} $60.08\%$) results while also improving Pedestrian detection performance.
In comparison to the baseline~\cite{yang2023bevheight}, the CFS fusion module constructs complementary BEV features through two-stage feature selections, thereby significantly improving detection results across all categories.

\begin{table*}[t]
\renewcommand{\arraystretch}{1.3}
    \begin{center}
        \caption{{\textbf{Ablations on hybrid-lifting BEV feature on the DAIR-V2X-I dataset}~\cite{yu2022dair}\textbf{.}}}
        \label{tab:bev-source}
        \resizebox{1.0\textwidth}{!}{
\setlength{\tabcolsep}{4mm}{   
\begin{tabular}{cccc|ccc|ccc|ccc}
\hline
 \multirow{2}{*}{\textbf{Depth-based}} & \multirow{2}{*}{\textbf{Height-based}} & \multicolumn{2}{c|}{\textbf{Implicit-based}} & \multicolumn{3}{c|}{\textbf{Vehicle $(IoU=0.5)$}} & \multicolumn{3}{c|}{\textbf{Pedestrian $(IoU=0.25)$}} & \multicolumn{3}{c}{\textbf{Cyclist $(IoU=0.25)$}} \\
& & Transformer & MLP & Easy & Middle & Hard & Easy & Middle & Hard & Easy & Middle & Hard \\
\hline
\hline

\checkmark & - & - & - & 73.05 & 61.32 & 61.19 & 22.10 & 21.57 & 21.11 & 42.85 & 42.26 & 42.09 \\   %
- & \checkmark & - & - & 76.61 & 64.71 & 64.76 & 27.34 & 26.09 & 26.33 & 49.68 & 48.84 & 48.58 \\   %
- & - & \checkmark & - & 61.37 & 50.73 & 50.73 & 16.89 & 15.82 & 15.95 & 22.16 & 22.13 & 22.06 \\   %
\rowcolor{gray!20}
\checkmark & \checkmark & - & - & \textbf{78.02} & \textbf{65.90} & \textbf{68.09} & \textbf{32.72} & \textbf{31.06} & \textbf{31.29} & \textbf{55.56} & \textbf{56.31} & \textbf{57.05} \\  %
\checkmark & \checkmark & \checkmark & - & 76.65 & 64.79 & 64.89 & 31.56 & 30.00 & 30.27 & 52.23 & 52.88 & 53.61 \\ %
\checkmark & \checkmark & - & \checkmark & 77.24 & 65.19 & 65.28 & 32.44 & 30.90 & 31.17 & 49.87 & 51.76 & 52.32 \\ %

\hline

\end{tabular}
}
}
    \end{center}
    \vskip -5ex
\end{table*}

\noindent\textbf{Hybrid-Lifting BEV Feature.} 
Tab.~\ref{tab:bev-source} shows our exploration of how heterogeneous BEV features reinforce each other. 
As for transformer-based BEV features, we employ the BEVFormer-style~\cite{li2022bevformer} using cross-attention mechanisms. 
For MLP-based BEV features, linear layers are used for mapping to achieve perspective transformation. 
Experimental results indicate that depth-based and height-based BEV features exhibit the best complementarity. 
Introducing additional transformer-based or MLP-based BEV features does not further enhance detection performance, aligning with our original motivation for CoBEV construction and observations of accuracy distributions across depth and height intervals.

\begin{table*}[t]
\renewcommand{\arraystretch}{1.3}
    \begin{center}
        \caption{\textbf{Ablations on point cloud supervision on the DAIR-V2X-I dataset}~\cite{yu2022dair}\textbf{.}}
        \label{tab:pcd-supervision}
        \resizebox{0.9\textwidth}{!}{
\setlength{\tabcolsep}{4mm}{ 
\begin{tabular}{cc|ccc|ccc|ccc}
\hline
 \multirow{2}{*}{\textbf{Depth-supervision}} & \multirow{2}{*}{\textbf{Height-supervision}} & \multicolumn{3}{c|}{\textbf{Vehicle $(IoU=0.5)$}} & \multicolumn{3}{c|}{\textbf{Pedestrian $(IoU=0.25)$}} & \multicolumn{3}{c}{\textbf{Cyclist $(IoU=0.25)$}} \\
& & Easy & Middle & Hard & Easy & Middle & Hard & Easy & Middle & Hard \\
\hline
\hline

Using GT & - & 82.27 & 70.02 & 70.06 & 64.19 & 61.83 & 62.05 & 68.88 & 69.63 & 70.09 \\ %
- & Using GT & 61.80 & 53.25 & 51.49 & 39.20 & 37.48 & 37.71 & 55.47 & 58.64 & 59.15 \\
Using GT & Using GT & 82.64 & 70.29 & 70.33 & 62.76 & 60.49 & 60.81 & 66.52 & 65.00 & 65.45 \\

\hline

\rowcolor{gray!20}
- & - & 81.20 & 68.86 & 68.99 & \textbf{44.23} & \textbf{42.31} & \textbf{42.55} & \textbf{61.28} & \textbf{61.00} & \textbf{61.61} \\
\checkmark & -  & \textbf{81.33} & \textbf{69.08} & \textbf{69.18} & 42.86 & 40.89 & 41.22 & 59.62 & 58.32 & 58.94 \\
- & \checkmark  & 81.27 & 69.01 & 69.10 & 42.81 & 40.79 & 41.07 & 59.85 & 57.29 & 57.88 \\
\checkmark & \checkmark  & 79.04 & 66.85 & 69.05 & 43.81 & 41.93 & 42.24 & 59.76 & 60.91 & 61.45 \\

\hline

\end{tabular}
}
}
    \end{center}
    \vskip -5ex
\end{table*}

\noindent\textbf{Point Cloud Supervision for Depth or Height.} 
Considering LiDAR's capacity to precisely measure the depth and height of roadside targets, we experiment with introducing point clouds for depth or height supervision. 
Surprisingly, contrary to prior work~\cite{li2022bevdepth}, we discover that refraining from introducing any form of point cloud supervision for depth and height effectively enhances detection accuracy across all categories. 
Introducing supervision solely for depth or height further improves vehicle detection performance but hinders the results on small targets. 
When both depth and height supervision are applied, vehicle detection accuracy drops significantly ($66.85\%$ \textit{vs.} $68.86\%$). 
Our findings corroborate that in BEVHeight~\cite{yang2023bevheight}, which observes similar trends for height supervision. 
To delve into this phenomenon, we replace the estimated depth/height distribution with depth or height ground truth during test time after training and observe corresponding accuracy changes. 
Depth ground truth significantly improves detection accuracy ($70.02\%$ \textit{vs.} $68.86\%$), while height ground truth weakens the detector's performance ($53.25\%$ \textit{vs.} $68.86\%$). 
We argue that this is because the height detector tends to learn the overall height interval of vehicles, pedestrians, and other categories, rather than specific height values for each pixel, focusing more on semantic cues during training than geometric ones. 
CoBEV leverages the rich semantic context of the height detector and the precise geometric cues of the depth detector, therefore showing the ability to construct robust BEV features.

\begin{table}[t]
\renewcommand{\arraystretch}{1.3}
    \begin{center}
        \caption{{\textbf{Ablations on knowledge distillation on the DAIR-V2X-I dataset}~\cite{yu2022dair}\textbf{.}}}
        \label{tab:kd}
        \resizebox{0.5\textwidth}{!}{ 
\setlength{\tabcolsep}{0.5mm}{ 
\begin{tabular}{lc|ccc|ccc|ccc}
\hline
\multirow{2}{*}{\textbf{Case}} & \multirow{2}{*}{\textbf{Modality}} & \multicolumn{3}{c|}{\textbf{Vehicle}} & \multicolumn{3}{c|}{\textbf{Pedestrian}} & \multicolumn{3}{c}{\textbf{Cyclist}} \\
& & Easy & Middle & Hard & Easy & Middle & Hard & Easy & Middle & Hard \\
\hline
\hline

Teacher & C+L & \textcolor{gray}{82.57} & \textcolor{gray}{70.28} & \textcolor{gray}{70.33} & \textcolor{gray}{73.01} & \textcolor{gray}{70.61} & \textcolor{gray}{70.66} & \textcolor{gray}{73.95} & \textcolor{gray}{76.31} & \textcolor{gray}{76.69} \\
Student & C & 81.20 & 68.86 & 68.99 & 44.23 & 42.31 & 42.55 & 61.28 & 61.00 & 61.61 \\

\hline

FitNet~\cite{romero2014fitnets} & C & 81.00 & 68.71 & 68.82 & 44.41 & 42.47 & 42.69 & 63.75 & 62.25 & 62.93 \\
CMKD~\cite{hong2022cross} & C & 81.14 & 68.80 & 68.90 & 44.53 & 42.51 & 42.84 & 62.19 & 63.53 & 64.17 \\
BEVDistill~\cite{chen2022bevdistill} & C & 81.38 & 69.05 & 69.18 & 45.61 & 43.50 & 43.93 & 63.63 & 64.40 & 65.00 \\
UniDistill~\cite{zhou2023unidistill} & C & 81.71 & 69.33 & 69.43 & 48.78 & 46.57 & 46.86 & 64.34 & 65.32 & 65.77 \\
\rowcolor{gray!20}
Ours & C & \textbf{82.01} & \textbf{69.57} & \textbf{69.66} & \textbf{49.32} & \textbf{47.21} & \textbf{47.48} & \textbf{66.13} & \textbf{66.17} & \textbf{66.69} \\
\textit{w.r.t. CoBEV} & - & \red{+0.81} & \red{+0.71} & \red{+0.67} & \red{+5.09} & \red{+4.90} & \red{+4.93} & \red{+4.85} & \red{+5.17} & \red{+5.08} \\

\hline

\end{tabular}
}
}
\begin{flushleft}
-- Note: `C': Camera is used, `C+L': both Camera and LiDAR are used.
\end{flushleft}
    \end{center}
    \vskip -5ex
\end{table}

\noindent\textbf{Knowledge Distillation.} 
In Tab.~\ref{tab:kd}, we compare the popular knowledge distillation frameworks~\cite{zhou2023unidistill,hong2022cross,romero2014fitnets,chen2022bevdistill} with the proposed BEV feature distillation scheme.
The teacher model, incorporating fused modalities, excels at handling small targets like pedestrians and cyclists. 
Consequently, the student model can more effectively emulate the teacher model's prior knowledge regarding small object detection problems. 
The proposed BEV feature distillation scheme adeptly handles erroneous supervision signals when the teacher model performs poorly, introducing the distill adapter at low-level features to enhance the detection accuracy of large targets like Vehicles (${+}0.71\%$). 
The student model closely replicates the teacher model at high-level BEV features, thus capturing the structural knowledge of the traffic scene independent of target size. 
This approach avoids suppressing visual cues of small targets, resulting in substantial improvements in the detection accuracy of Pedestrians and Cyclists (${+}4.90\%$ / ${+}5.17\%$), outperforming existing distillation methods.

\section{Conclusion}

In this paper, we introduce CoBEV, a monocular camera-based 3D detector tailored for roadside scenes, leveraging complementary depth and height information to create robust BEV representations that enhance traffic scene understanding. CoBEV achieves state-of-the-art accuracy on DAIR-V2X-I and Rope3D datasets by integrating precise geometric cues from depth features with distinct semantic contexts from height features. It demonstrates exceptional adaptability to different cameras and intersection scenes, and shows enhanced resilience to camera parameter variations through its robust BEV features from a complementary feature selection module. Moreover, we detail a BEV feature distillation method designed for roadside scenarios, enhancing the detector's understanding of scene semantics and structural relationships, thereby improving size-agnostic detection accuracy. Performance analysis across various depth and camera installation heights confirms the effective interplay between depth and height detectors. Extensive experiments and ablation studies verify CoBEV's effectiveness.
{Additionally, CoBEV's qualitative testing under diverse weather conditions underscores its strong generalizability. We emphasize the future need for more varied roadside datasets to ensure comprehensive testing and reliability of detection methods in different scenarios.}

{\small
\bibliographystyle{IEEEtran}
\bibliography{bib}
}

\appendices
\counterwithin{figure}{section}
\counterwithin{equation}{section}

\section{Network Overview}
\label{sec:network_overview}
As shown in Fig.~5, given a monocular roadside image frame $X {\in} \mathbb{R} ^ {3 \times H \times W}$ with its corresponding extrinsic matrix $E {\in} \mathbb{R} ^ {3 \times 4}$ and intrinsic matrix $I {\in} \mathbb{R} ^ {3 \times 3}$. Formally, a 2D convolutional image encoder with FPN neck maps $X$ to the high-dimensional image feature $F {\in} \mathbb{R} ^ {C \times \frac{H}{16} \times \frac{W}{16}}$. Then, $F$ and the camera parameters $ \{E, I \} $ are fed into the \textit{Camera-aware Hybrid Lifting} module.
The purpose of this stage is to lift the monocular image features $F$ from the 2D coordinate system to the 3D frame by calculating the depth distribution, feature context, and height distribution, respectively.
The partial-pillar voxel pooling module then compresses the 3D features from the pillars on the BEV plane, which are voxels with height.
Note that we perform height compression instead of completely flattening the features to a 2D plane in order to preserve a certain height axis.
Therefore, the free flow of relevant visual cues in all dimensions can be achieved in the next novel \textit{Complementary Feature Selection (CFS)} module, and then construct complementary BEV feature $F_{com} {\in} \mathbb{R} ^ {C_{com} \times \frac{H}{16} \times \frac{W}{16}}$.
Finally, the complementary BEV feature $F_{com}$ is fed into the detection head, 
in which the 3D bounding boxes $\mathbf{B_{ego}}$ composed of position $(x,y,z)$, dimension $(l,w,h)$, and orientation $ \theta $ is output. 
Moreover, we design 
a fusion-to-camera \textit{BEV Feature Distillation} framework by using a teacher that fuses the multi-modality information of LiDAR and camera to further enhance the accuracy of monocular 3D detection agnostic to the target size.

\section{Training Objectives}
\label{sec:training_objectives}
The final loss function can be expressed as:
\begin{equation}
\label{equ:loss_final}
      \mathcal{L}_{total} = \mathcal{L}_{det} + \mathcal{L}_{low} + \mathcal{L}_{high} + \mathcal{L}_{res}, 
\end{equation}
where the detection loss $\mathcal{L}_{det}$ inherits from~\cite{yan2018second}.

\section{Implementation Details}
Unless otherwise specified, the CoBEV detector employs ResNet101~\cite{he2016deep} as its image encoder, while ResNet50~\cite{he2016deep} is used for ablation experiments.
The input image size is set to $864 {\times} 1536$.
The width range of the Bird's Eye View (BEV) grid spans across $ \lbrack {-}51.2m, 51.2m \rbrack$, 
with a length range of $ \lbrack 0m, 102.4m \rbrack$. 
 {The cell size of the BEV grid is $0.8m {\times} 0.8m$ for CoBEV~(ResNet50) and CoBEV*~(ResNet101),  $0.4m {\times} 0.4m$ for CoBEV (ResNet101) and CoBEV$_{full}$~(ResNet101).}
The depth range $D$ of the target is defined as $ \lbrack 2m, 104.4m \rbrack$, with a unit size of $0.4m$, employing a Uniform Discretization (UD) that yields a total of $256$ bins. 
The height range $H$ of the target varies depending on the statistical distribution of the specific dataset.
For DAIR-V2X-I~\cite{yu2022dair} and Rope3D~\cite{ye2022rope3d} datasets, we adhere to the official definitions, with height ranges of $ \lbrack {-}2m, 0m \rbrack$ and $ \lbrack {-}1.5m, 3m \rbrack$, respectively.
In the case of the Supremind-Road dataset, the height range spans from $ \lbrack {-}1m, 4m \rbrack$.
The height cell size follows dynamic partitioning as outlined in Equ.~\ref{equ:cam_mlp}
with $\alpha$ set to $1.5$.
CoBEV is implemented in PyTorch and incorporates random 2D rotation and scaling augmentation. Training is performed individually on all datasets, with a batch size of $16$ for $150$ epochs on each dataset. 
The learning rate is set to $2e{-}4$, and the optimizer is AdamW~\cite{loshchilov2017decoupled}.

\section{Evaluation Metrics}
Following the established detection metrics employed in various benchmark datasets, our evaluation differs depending on the dataset.
For the DAIR-V2X-I dataset~\cite{dai2021attentional}, we report the $40$-point average precision ($AP_{3D|R40}$) of bounding boxes~\cite{geiger2012we}, which is further categorized into three modes: easy, middle, and hard, based on the box characteristics such as size, visibility, and truncation~\cite{everingham2010pascal}.
In the case of the Rope3D dataset~\cite{ye2022rope3d}, we employ the $AP_{3D|R40}$ and the $Rope_{score}$~\cite{ye2022rope3d} for assessment.
The $Rope_{score}$ provides a comprehensive evaluation, taking into account factors such as bounding box center, orientation, area, and ground points. We present results for Rope3D under two conditions: easy and strict, characterized by bounding box Intersection-over-Union (IoU) thresholds of $0.5$ and $0.7$, respectively.
For the Supremind-Road dataset, we utilize $AP_{3D|R40}$ as the evaluation metric, and categorize it into easy and hard modes, with the distinction based on the truncation of the boxes. According to different target sizes, the IoU threshold is $0.5$ for vehicles and $0.25$ for pedestrians, cyclists, and tricycles.

\section{Future Work}
In the future, we are interested in exploring the performance of the CoBEV framework in vehicle-side multi-camera 3D detection tasks, aiming for a unified, high-precision, and robust detector for both vehicles and infrastructure.
Moreover, we intend to leverage generative techniques, such as diffusion models, to simulate and augment training data, particularly in scenarios with limited real-world samples, like traffic accident scenarios.
This approach will enhance the model's reliability under safety-critical conditions and help to address the long tail problem in training.
We are also keen to explore the application of the CoBEV framework in addressing perception challenges related to vehicle-to-road collaboration and multi-modal fusion.
Our goal is to enhance the perception range of intelligent vehicles and improve the responsiveness of the overall transportation system by constructing robust features within a unified BEV space.

\section{Qualitative Result Visualizations}
 {In this section, we further present qualitative comparison results between BEVDepth~\cite{li2022bevdepth}, BEVHeight~\cite{yang2023bevheight}, and the proposed CoBEV across the Rope3D~\cite{ye2022rope3d} and Supremind-Road datasets. Furthermore, in Fig.~\ref{figure:robust_comp}, we provide a qualitative comparison of the robustness of the CoBEV framework in comparison to previous methods when subjected to camera parameter disturbances with the BEV view of LiDAR.}

\begin{figure*}[!t]
\centering
\includegraphics[width=1.0\linewidth]{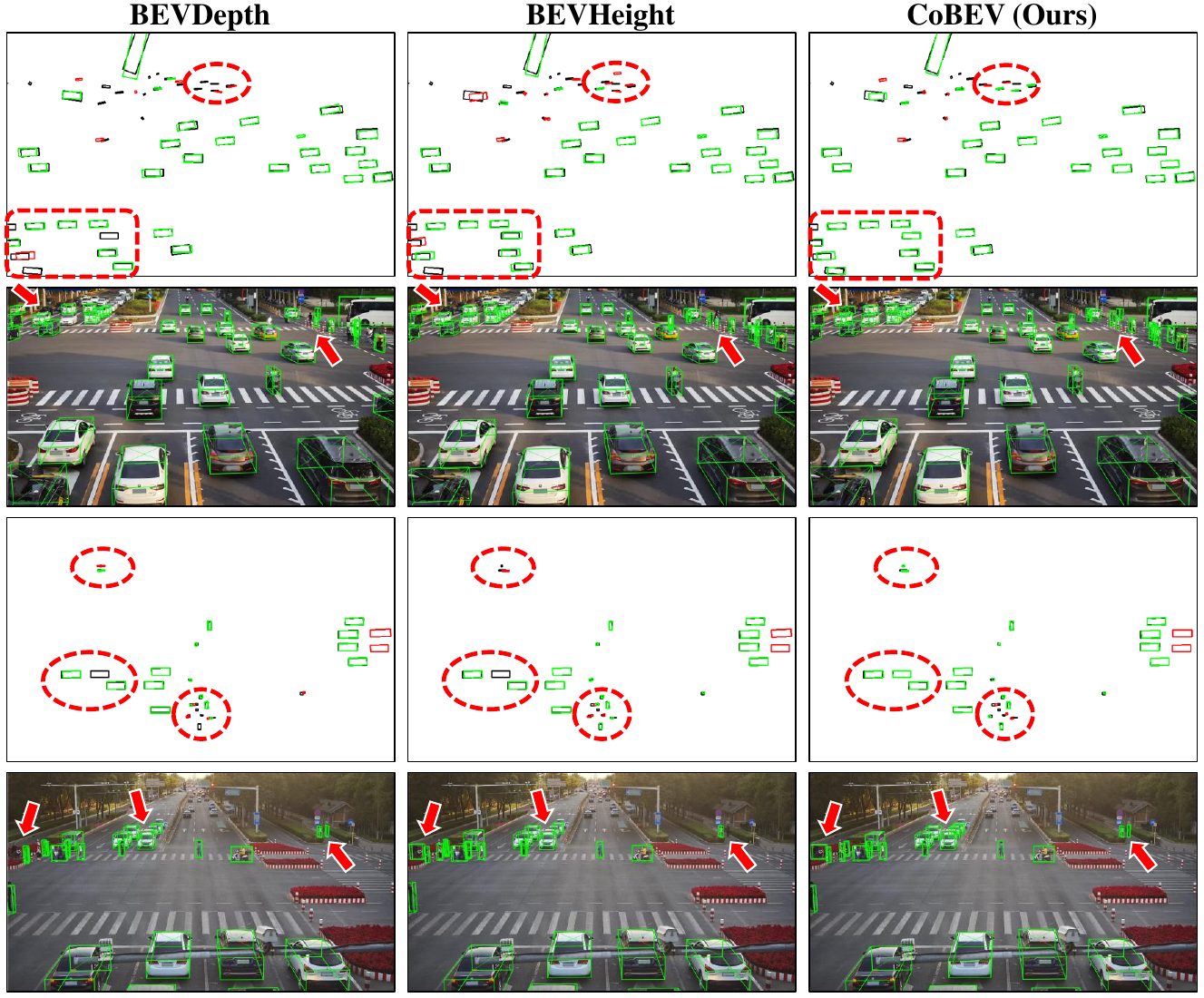}
\caption{\textit{Qualitatively Comparisons on the Rope3D dataset.} BEVDepth and BEVHeight prove less effective in handling small targets like pedestrians and cyclists. CoBEV, on the other hand, enhances the detection accuracy of such small targets by constructing complementary, fine-grained BEV representations.
}
\label{figure:rope_comp}
\end{figure*}

\begin{figure*}[!t]
\centering
\includegraphics[width=1.0\linewidth]{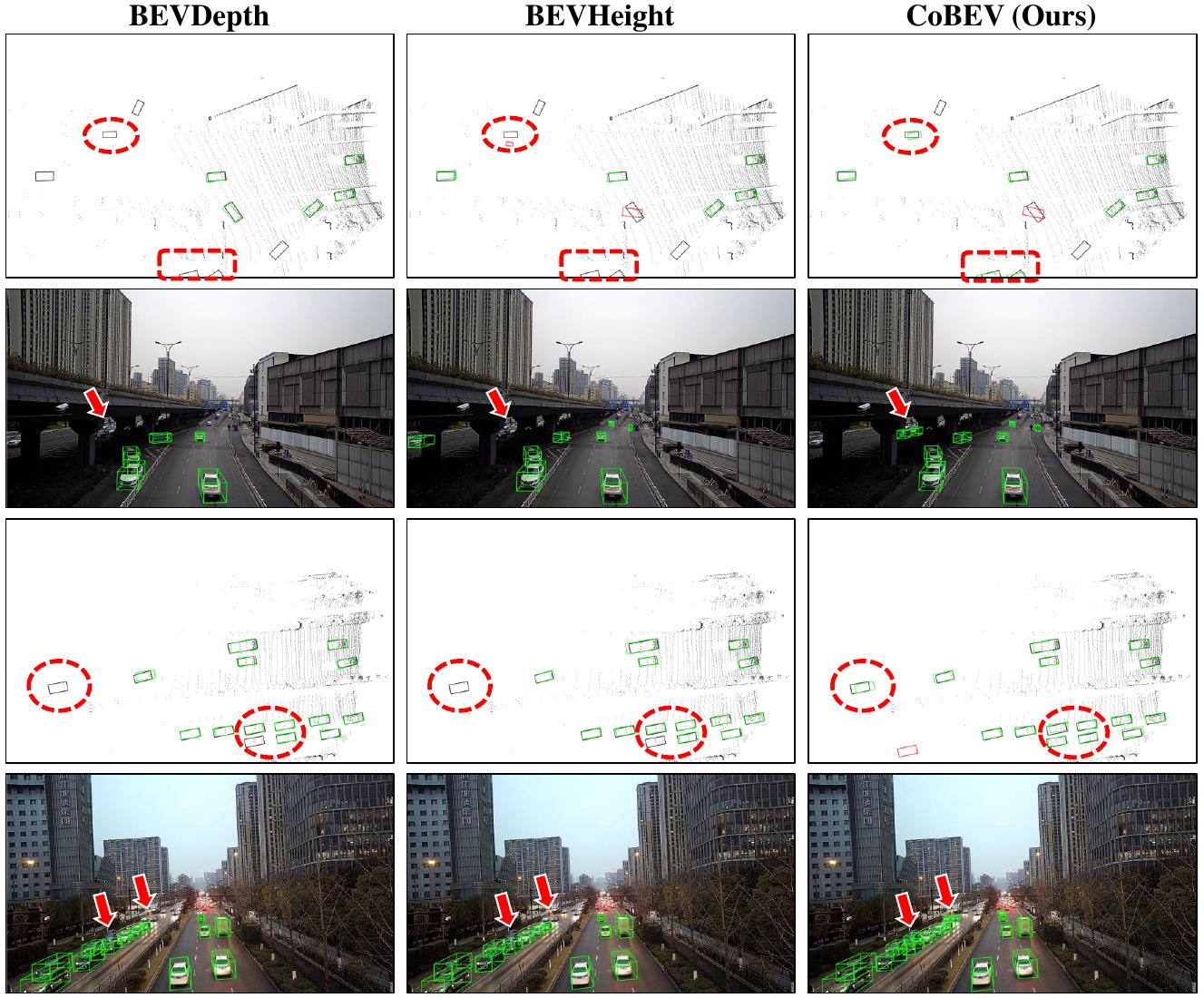}
\caption{\textit{Qualitatively Comparisons on the Supremind-Road dataset.} 
In comparison to BEVDepth and BEVHeight, the proposed CoBEV mitigates false alarms and missed detections, simultaneously enhancing the accuracy of detecting challenging long-distance targets.
}
\label{figure:sm_comp}
\end{figure*}

\begin{figure*}[!t]
\centering
\includegraphics[width=1.0\linewidth]{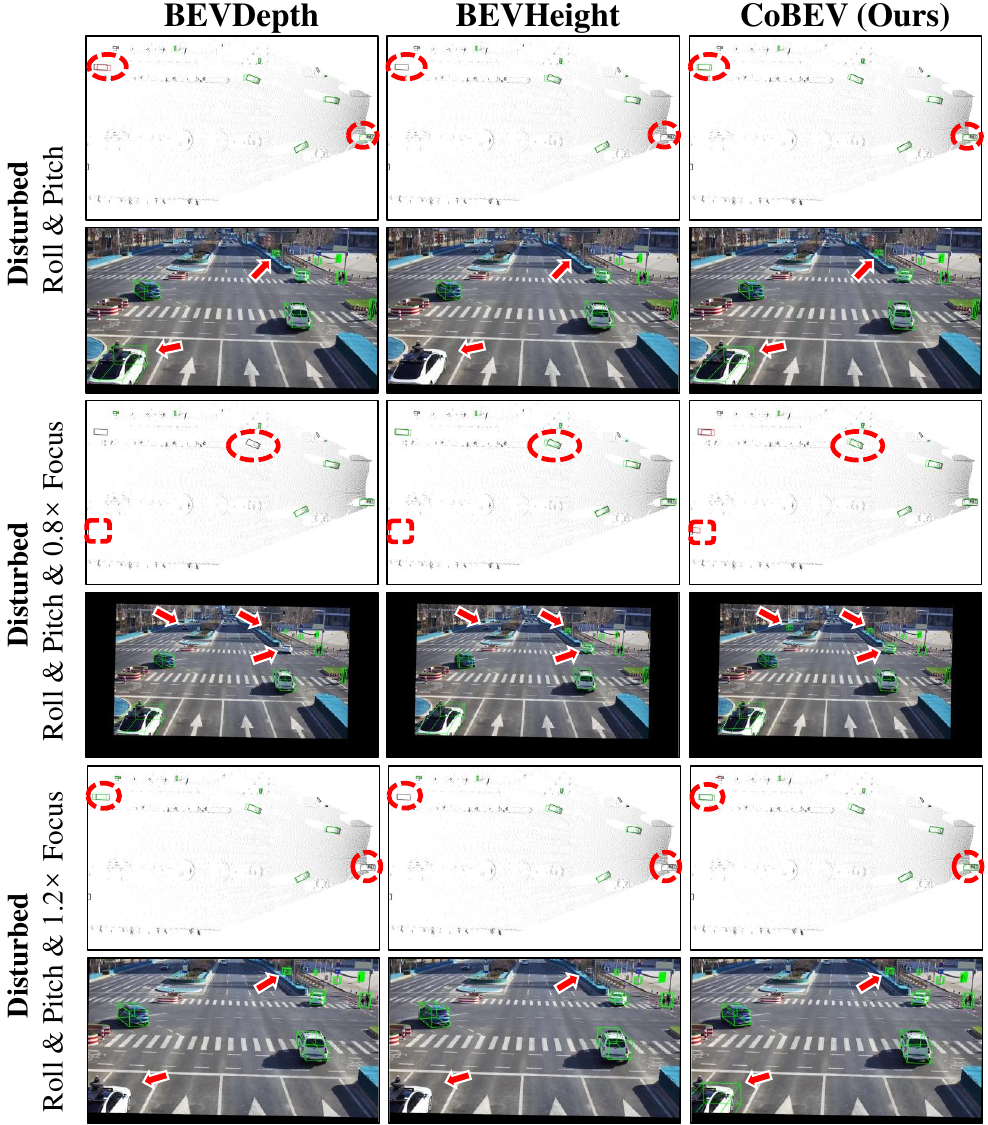}
\caption{ {\textit{Qualitatively comparisons on the DAIR-V2X-I dataset under the camera parameters disturbances with BEV view of LiDAR.}} In the condition of various camera parameter disturbances, including focal length, roll, and pitch, CoBEV consistently demonstrates the most robust target detection capabilities. This applies to both targets truncated at close range and small targets situated at long distances.
}
\label{figure:robust_comp}
\end{figure*}

\end{document}